\documentclass[10pt,journal,compsoc]{IEEEtran}
\ifCLASSOPTIONcompsoc
  \usepackage[nocompress]{cite}
\else
  \usepackage{cite}
\fi
\ifCLASSINFOpdf
\else
\fi
\usepackage{tikz}
\usepackage{comment}
\usepackage{amsmath,amssymb} 
\usepackage{color}
\usepackage{array}
\usepackage{float}
\usepackage{tabularx}
\usepackage{multicol}
\usepackage{booktabs}
\usepackage{colortbl}
\usepackage{amssymb}
\usepackage{pifont}
\usepackage{tikz}
\usepackage{textcomp}
\usepackage{gensymb}
\usepackage{adjustbox}
\usepackage{bigstrut, multirow}
\usepackage{arydshln}
\usepackage[hidelinks]{hyperref}

\newcommand{\etal}{\textit{et al}. }
\newcommand{\eg}{\textit{e.g.} }
\newcommand{\ie}{\textit{i.e.} }

\usepackage{dsfont}

\newif\ifshowedits

\newcommand{\addeditor}[3]{%
  \definecolor{#1color}{rgb}{#3}
  \expandafter\newcommand\csname #1\endcsname[1]{
  \ifshowedits
    {\color{#1color} ##1}
  \else
    {##1}
  \fi
  }%
  \expandafter\newcommand\csname #1rmk\endcsname[1]{
  \ifshowedits
    {\color{#1color} {\bf [#2: ##1]}}
  \else
    {}
  \fi
  }%
}

\newcommand{\newtvar}[1]{
  \expandafter\newcommand\csname #1\endcsname{\text{#1}}
}

\newcommand{\calC}{{\cal C}}
\newcommand{\calD}{{\cal D}}

\newcommand{\calL}{{\cal L}}

\newcommand{\calP}{{\cal P}}

\DeclareMathOperator*{\argmin}{arg\,min}

\newcommand{\Ls}{\calL_s}
\newcommand{\Lreg}{\calL_\text{reg}}

\newcommand{\Ldepth}{\calL_\text{3D}}
\newcommand{\Lemb}{\calL_\text{2D}}

\addeditor{sinisa}{SS}{1.0, 0.0, 0.0}
\addeditor{alireza}{AM}{0.0, 0.0, 0.8}
\addeditor{friedrich}{FF}{0.0, 0.0, 0.5}
\addeditor{vincent}{VL}{0.5, 0.5, 0.0}

\begin{document}
%
\title{MCTS with Refinement for Proposals Selection Games in Scene Understanding}

%
%
%
%

\author{Sinisa~Stekovic, Mahdi~Rad, Alireza~Moradi, Friedrich~Fraundorfer, and Vincent~Lepetit
\IEEEcompsocitemizethanks{\IEEEcompsocthanksitem S.~Stekovic, M.~Rad, A.~Moradi, F.~Fraundorfer and V.~Lepetit are with the Institute for
Computer Graphics and Vision, Graz University of Technology, Graz,
Austria. \protect\\
\IEEEcompsocthanksitem V.~Lepetit is also with Universit\'e Paris-Est, \'Ecole des Ponts ParisTech, Paris, France. \\ 
E-mail: sinisa.stekovic@icg.tugraz.at, rad@icg.tugraz.at, alirezamoradi1044@gmail.com, fraundorfer@icg.tugraz.at, \\ vincent.lepetit@enpc.fr \protect\\
\IEEEcompsocthanksitem {\small Project page: \href{https://www.tugraz.at/index.php?id=52770}{ \color{black} https://www.tugraz.at/index.php?id=52770}}}
\thanks{Manuscript received March 31, 2022;}}

%
%

\markboth{IEEE TRANSACTIONS ON PATTERN ANALYSIS AND MACHINE INTELLIGENCE}%
{Stekovic \MakeLowercase{\textit{et al.}}: MCTS with Refinement for Proposals Selection Games in Scene Understanding}
%



\IEEEtitleabstractindextext{%
\begin{abstract}
We propose a novel method applicable in many scene understanding problems that adapts the Monte Carlo Tree Search~(MCTS) algorithm, originally designed to learn to play games of high-state complexity. From a generated pool of proposals, our method jointly selects and optimizes proposals that minimize the objective term. In our first application for floor plan reconstruction from point clouds, our method selects and refines the room proposals, modelled as 2D polygons, by optimizing on an objective function combining the fitness as predicted by a deep network and regularizing terms on the room shapes. We also introduce a novel differentiable method for rendering the polygonal shapes of these proposals. Our evaluations on the recent and challenging Structured3D and Floor-SP datasets show significant improvements over the state-of-the-art, without imposing hard constraints nor assumptions on the floor plan configurations. In our second application, we extend our approach to reconstruct general 3D room layouts from a color image and obtain accurate room layouts. We also show that our differentiable renderer can easily be extended for rendering 3D planar polygons and polygon embeddings. Our method shows high performance on the Matterport3D-Layout dataset, without introducing hard constraints on room layout configurations.



\end{abstract}

}

\maketitle

\IEEEdisplaynontitleabstractindextext

%
\IEEEpeerreviewmaketitle

\IEEEraisesectionheading{\section{Introduction}\label{sec:introduction}}

%
%
%
%

\IEEEPARstart{S}{cene} understanding from images is one of the main topics in computer vision, as it aims both at replicating one of the key abilities of human beings and producing solutions for many applications such as robotics or augmented reality. We focus in this work on tasks that are relevant in for the field of Architecture, Engineering and Construction~(AEC). Namely, we consider two tasks where we achieve state-of-the-art performance: floor plan reconstruction, where each room of an indoor environment is represented as a polygon with one edge per wall, and 3D room layout reconstruction, where each wall of a single room is modelled as a 3D planar polygon. This work extends  our original work~\cite{stekovic2021montefloor} on floor plan estimation with additional experiments that show generalization of our method to the 3D room layout estimation task.

\begin{figure}
    \centering
    \begin{tabular}{ccc}
        \includegraphics[width=0.35\linewidth]{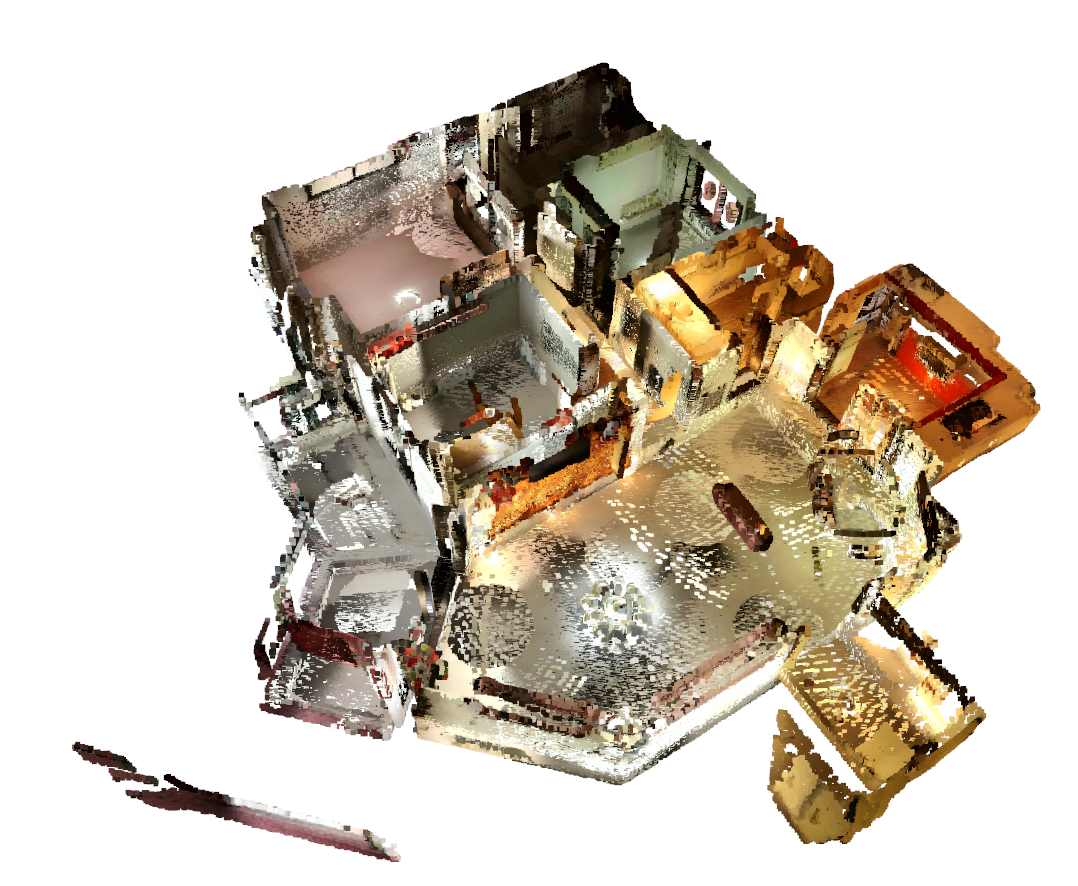} &  
        \includegraphics[width=0.25\linewidth]{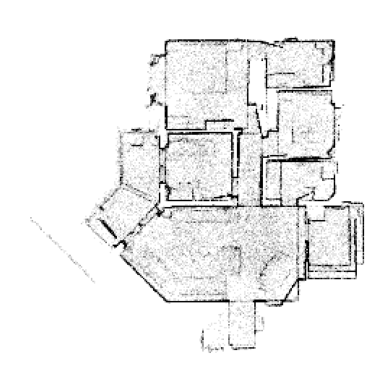} &
        \includegraphics[width=0.25\linewidth]{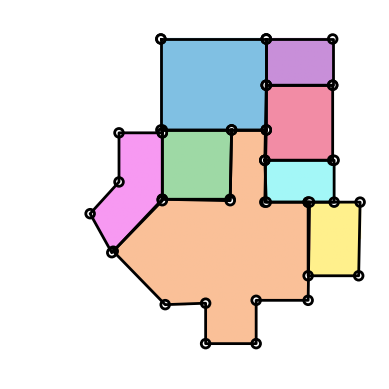} \\
        Point cloud & Density map & Our result \\
    \end{tabular}
    \begin{tabular}{cc}
        \includegraphics[width=0.25\linewidth]{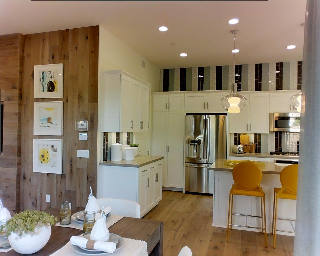} & \includegraphics[width=0.25\linewidth]{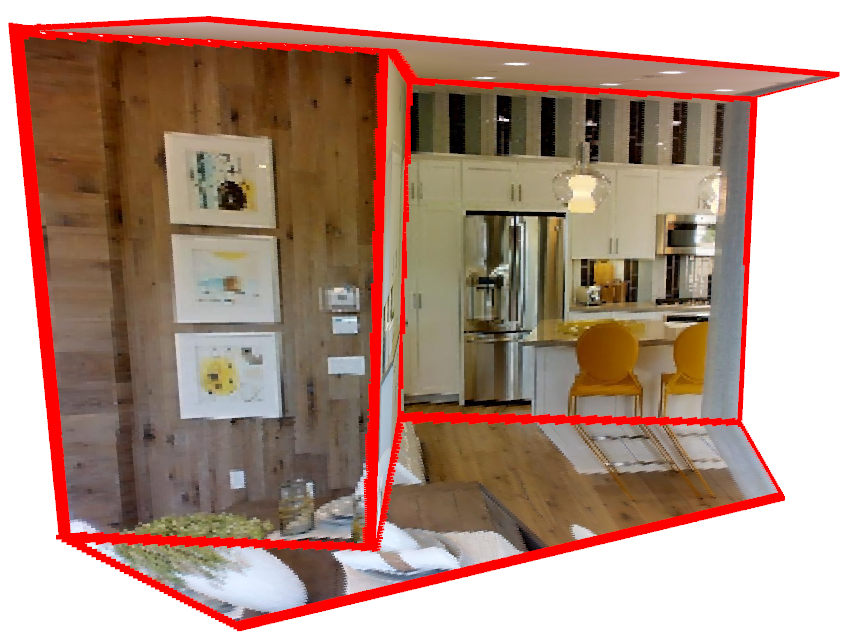} \\
        Color image & Our result \\
    \end{tabular}
    \vspace{-0.3cm}
    \caption{Our method is applicable to different tasks in scene understanding. In the first row, given a density map \textit{i.e.} the top view of the 3D point cloud of a floor, we retrieve an accurate floor map that successfully recovers a variety of room shapes. In the second row, we accurately reconstruct general 3D room layouts from a single color image.}
    \label{fig:teaser}
\end{figure}

While we focus in this work on floor plan reconstruction and room layout reconstruction, our approach is general and is applicable to other tasks in scene understanding. We formulate a "proposals selection game" as a family of tasks in scene understanding where our approach can be applied. Our main component is generic: We start from proposals for the target objects~(the rooms or the walls in our applications). This step does not have to perform well to obtain good final results as our Monte Carlo Tree Search~(MCTS)-based algorithm can deal with many false positives. This algorithm looks for the final solution by maximizing a data-driven score, which can thus be easily replaced to adapt to another problem. During the search our algorithm refines sampled solutions hence performing the discrete and continuous optimization jointly.

In our main application we focus on reconstructing floor plans from unstructured 3D point clouds as in~\cite{avetisyan2020scenecad,chen2019floor,Liu2018floornet,murali2017indoor}, as they can now be generated easily with an RGB-D camera and can cover an entire floor. To estimate the floor plan from a given point cloud, \cite{chen2019floor,Liu2018floornet} proposes to first project the point cloud into a virtual top view to create a 'density map', as the walls, the main features for creating the floor map, appear relatively clearly in the density map. As shown in Figure~\ref{fig:teaser}, the density maps can be noisy, and it is still challenging to represent the rooms as vector drawings with a minimal number of edges as a human designer would do, especially for non-Manhattan floor maps. To deal with this complexity, Floor-SP~\cite{chen2019floor} proposes a graph-based solution with a sound energy term but still assumes the existence of some dominant wall directions in the scene.

Our main contribution is a method, which we call MonteFloor, that is conceptually simple and robust and  returns high-quality floor plans. Figure~\ref{fig:teaser} shows an example from the Floor-SP test set that demonstrates we can reconstruct complex floor maps, including very large ones with complex room shapes without having to tune hyper-parameters.

Like Floor-SP\cite{chen2019floor}, our method starts from room proposals generated by Mask-RCNN~\cite{he2017mask} from the density map. However, the way we handle these room proposals is fundamentally different from \cite{chen2019floor}. Where \cite{chen2019floor} adjusts the room walls and corners in a greedy fashion, we select the correct room proposals \emph{jointly} \emph{while adjusting their locations and shapes}, guided by a learned scoring function. 

This is possible thanks to two main contributions. Our first contribution is based on the Monte Carlo Tree Search~(MCTS) algorithm~\cite{browne2012survey,Coulom06}. MCTS is a stochastic algorithm that efficiently explores search trees and has been used for example in AlphaGo and AlphaZero to select moves when playing Go or other games with high combinatorials~\cite{AlphaZero}. We use it to search among the room proposals for the ones actually belonging to the correct floor plan. In our case, a move corresponds to the selection of a room proposal. In contrast with other tree search algorithms, MCTS is based only on the evaluations of leaves. This means that we can select a set of proposals based on how well they explain the density map \emph{together}.  After evaluating a leaf, MCTS updates a score in the visited nodes, which will be used to guide the next tree explorations. 

To evaluate how well a set of proposals explains an input density map, we introduce an objective term that combines a 'metric score' predicted by a deep network and regularization terms.  This network takes as input the density map and an image of the selected proposals to predict the Intersection-over-Union between the selected proposals and the ground truth. The regularization terms encourage the selected room proposals to be in contact with each other without overlapping, and angles close to 90$\degree$ to be exactly 90$\degree$---note this is different from enforcing Manhattan World conditions as other angles are also accepted.

Moreover, to adapt MCTS and obtain accurate floor plan estimates, we extended it by adding a refinement step before evaluating the objective function.  The step performs an optimization of the objective function and adjusts the shapes of the selected room proposals to better fit the density map. This is made possible by our second contribution, which is a novel differentiable method to optimize the shapes of 2D polygons.  Note that very recently, \cite{MCSS} has also used MCTS for a scene understanding problem. However, it proposes a straightforward application of MCTS. By contrast, we rely on a learned objective function suitable to our problem, and we introduce an optimization step to obtain accurate estimates.

To evaluate our method and compare it with Floor-SP~\cite{chen2019floor}, which is the state-of-the-art for our problem, we first perform experiments on the Structured3D dataset~\cite{Structured3D} that contains a variety of complex layout configurations. We show significant improvements  regarding both the accuracy and time complexity over Floor-SP (after retraining their method on Structured3D). As the authors of Floor-SP could not provide the training set for their method, we could not re-train our network for predicting the metric specifically for this dataset, and we had to use the one trained on Structured3D. Despite this domain gap, we achieve better performance on the Floor-SP test set without imposing any hard constraints nor assumptions on the floor plan configurations.

\textbf{Extending MCTS with refinement to 3D room layout reconstruction.} To demonstrate the generalization aspect of our method, we additionally implement our MCTS with refinement scheme for the 3D room layout reconstruction task where the goal is to retrieve correct extents of layout components, such as walls, floors, and ceilings, from an input color image. While many works assume cuboid constraints~\cite{Hedau09,Lee2017roomnet,Hirzer2019smart}, we build our approach upon Stekovic~\etal~\cite{stekovic2020general} that only assumes planarity of layout components. As in \cite{stekovic2020general}, we start by generating a large pool of planar polygon proposals, and formulate a discrete optimization problem that looks for a subset in this pool that minimizes our objective term. Instead of simply performing an exhaustive search, we apply our MCTS with refinement scheme to reconstruct room layouts of high quality and reduce search and refinement efforts. We call the resulting method MonteRoom. Figure~\ref{fig:teaser} shows a challenging example where our method reconstructs very accurate 3D room layout. We perform evaluation on the Matterport3D-Layout dataset~\cite{zhang2020geolayout} that contains a range of complex room layout configurations and demonstrate that our method outperforms the current state-of-the-art method GeoLayout~\cite{zhang2020geolayout} by a significant margin.
\section{Related Work}
\label{sec:related_work}

\subsection{Floor Plan Estimation}

Early methods for floor plan creation from 3D data relied on basic image processing methods such as histograms or plane fitting~\cite{ff:huber20113dreconstruction,budroni2010automated,ff:huber2009automatedmodeling,sanchez2012planar,xiao2014reconstructing,xiong2013automatic}. For example, \cite{ff:huber2009automatedmodeling} creates a floor plan by detecting vertical planes in a 3D point cloud by building a histogram of the vertical positions of all measured points. In a similar way, \cite{budroni2010automated} creates a floor plan from located walls in a 3D point cloud by extracting planar structures by applying sweeping techniques to identify Manhattan-World directions. However, these techniques relied heavily on heuristics and were prone to fail on noisy data.

Significant progress has been made later by using graphical models as in \cite{ff:cabral2014planarfloorplan,furukawa2009reconstructing,gao2016multi,gao2014jigsaw,ff:ikehata2015structuredindoor}.  \cite{furukawa2009reconstructing} uses graph-cuts optimization in a volumetric MRF formulation. However, the proposed method is vulnerable to noisy data, as regularization in MRFs is based only on pairwise interaction terms. \cite{ff:ikehata2015structuredindoor} combines an MRF with Robust Principal Component Analysis to obtain  more compact 3D models. 
Graphical models are also used in \cite{gao2014jigsaw} where layouts and floor plans are recovered from crowd-sourced image and location data. 

Graph-based methods define objective functions made of unary terms representing the elements of the plan and binary terms which involve only two elements at a time (here, the elements are mostly walls). In our case, we use MCTS as the optimization algorithm. MCTS does not impose restrictions on the form of the objective function and we use an objective function that captures complex constraints. In particular, the main term of our objective function is a deep network that considers all the elements at the same time. Moreover, we complement MCTS by adding a refinement step to adjust the locations of the elements based on the same objective function.

More recent works rely on other optimization techniques~\cite{ff:Chao2013layoutestimation,chen2019floor,Liu2018floornet}. The challenges for these techniques, however, are the definition of a cost function and the optimization procedure. One of these methods called FloorNet~\cite{Liu2018floornet} proposes a deep network for detecting probable corner locations from a given density map of the scene, followed by an Integer Programming formulation. However,  incorrect corner detection and misdetections result in missing or extra walls and rooms. Moreover, the solution space is restricted to Manhattan scenes and generalizing to non-Manhattan scenes would lead to a much larger solution space. By contrast, our approach is scalable, as it relies on the efficiency of MCTS to reduce the search space, and can consider Manhattan  and non-Manhattan scenes with the same complexity. It selects room detections that best explains the input through a global optimization, and is thus not sensitive to false positives.

The starting point of our method is inspired by Floor-SP~\cite{chen2019floor}, which proposes to first segment room instances, and then to reconstruct polygonal representations of rooms by sequentially solving shortest path problems. In their case, every pixel location in a discretized density map is a node in a graph that potentially belongs to the polygonal curve of the room. Wrong segmentations may still lead to an inaccurate floor plan structure, while we handle incorrect room segmentation at an early stage.  Also, Floor-SP discretizes the edge directions of rooms and models multiple Manhattan frames per room, while our approach can consider any angle. It still encourages angles close to 90$\degree$ to be exactly 90$\degree$, which results in better shapes when the rooms actually follow Manhattan structures while allowing other shapes. As we will show in the experiments, our approach outperforms the accuracy of Floor-SP.

\subsection{Room Layout Reconstruction from Single Views}

More traditional approaches in room layout reconstruction relied on estimation of vanishing points ~\cite{Hedau09,Schwing12,Ramalingam13,Mallya15} to reconstruct room layouts under strong Manhattan-world assumptions. These approaches typically failed in scenarios where edge features of the layout are not visible in the image. Another track of room layout estimation methods tries to fit a single box that best describes the input scene~\cite{Schwing12, Zhang2013estimating,Mallya15,Lee2017roomnet,Zhang19,Hirzer2019smart}, some of which also include objects and humans in the room layout estimation loop~\cite{Holistic18,Chen2019holistic}.

Our 3D room layout reconstruction approach is more related to approaches that identify planes corresponding to the layout components. Zou~\etal~\cite{Zou2019complete} detects layout proposals with up-to-five dominant layout planes and applies a greedy hill-climbing algorithm to select among proposals based on the input RGB-D image. Recent data driven approaches for plane detection~\cite{liu2018planenet,Liu2019planercnn,yu2019single,tan2021planetr} became a very attractive option to extract general planes in the scene from a single color image. Howard-Jenkins~\etal~\cite{Howard2018outsidethebox} directly learns to estimate layout planes from an input color image and, similarly, we show in this work that we can train the PlaneTR~\cite{tan2021planetr} network to infer room layout planes. We build upon Stekovic~\etal~\cite{stekovic2020general} that, first, generates a set of room layout proposals by intersecting the detected layout planes, and, then, formulates a discrete optimization setting to find the optimal set of proposals. Instead of performing the brute force search, we rely on MCTS to find the optimal solution while jointly refining sampled solutions.

\subsection{Differentiable Rendering} 

Some works in 3D computer vision have shown interest in differentiable rendering~\cite{genova2020local,gkioxari2019mesh,groueix2018papier,mildenhall2020nerf,ravi2020pytorch3d,wang2018pixel2mesh}. However, these methods are focused on the rendering of 3D representations such as point clouds, voxels, meshes, and implicit 3D representations. In contrast, in this work we focus on fast differentiable rendering of 2D representations, \ie polygons, and introduce a differentiable winding algorithm for rasterization purposes. In our MonteRoom application we also show that our renderer can be easily extended to render polygonal embeddings and 3D planar polygons.

\section{Proposals Selection Games in Scene Understanding}

\begin{figure*}
    \centering
    \begin{tabular}{cc}
         \includegraphics[width=0.8\textwidth]{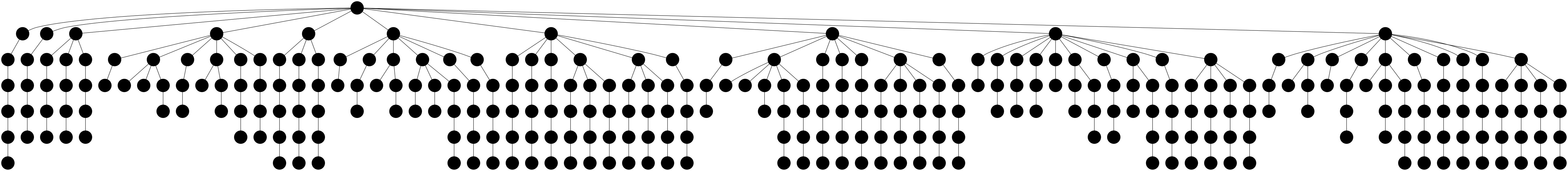} &  
         \includegraphics[width=0.1\textwidth]{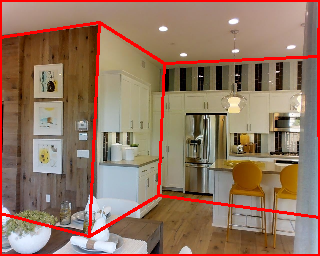} \\
         Solution tree built by MCTS & The best solution  
    \end{tabular}

    \caption{Visualization of a tree structure that our approach builds during search, in this example for a scene in our MonteRoom application. In our MonteFloor application the tree is often even deeper with larger amount of children nodes at lower levels of the tree. For each of the parent nodes, we only show $10$ children nodes with highest scores for visualization purposes. Our approach is able to focus on the most promising solutions even in such complex settings. It finds the best solution efficiently while, at the same time, refining sampled solutions in the search space. In this example, the best solution corresponds to the right-most path in the tree, and is visualized as a room layout overlay on top of the input image.}
    \label{fig:mcts_tree}
\end{figure*}

As we draw inspiration from successes in domain of high-state-complexity games, such as AlphaZero~\cite{AlphaZero}, we argue that it is possible to represent many problems in Scene Understanding as a \textit{proposals selection game}. 

It was, indeed, demonstrated in Monte Carlo Scene Search~(MCSS)~\cite{MCSS} that many methods, even state-of-the-art methods that rely on deep learning, still generate inaccurate solutions that contain some or many false positives. Our proposals selection game describes a general setting where we aim to recover true positives from a larger set of proposals.

Proposals selection game is a very simple and a very general single-player game. We initialize our game by generating a large set of proposals $\calP_0$ and the goal of the game is to retrieve a subset $\calP$ which best describes the input scene. The pool might contain empty proposals which are synthetically added for better handling of false positive proposals as we explain in Section~\ref{sec:mcts_ref}. Then, a move in the game consists of selecting one proposal from the pool. As the rule of scene understanding dictates that some proposals are mutually incompatible, every time we pop one proposal from the pool, we remove all incompatible proposals from the pool as well. The game continues until the pool of proposals becomes empty, after which the game ends and the final \textit{score} is calculated that represents the fitness of the selected solution to the given scene.

For example, we could design a CAD-model retrieval game as in MCSS~\cite{MCSS} where the pool of proposals contains CAD-models of furniture and structural elements, and the goal is to retrieve a subset of this pool that best fits the scene in 2D and 3D. Similarly, in this work we explore two further applications relevant to the field of
Architecture, Engineering and Construction~(AEC), namely, floor-plan reconstruction from top-view density maps and single image room layout estimation as we explain further in Section~\ref{sec:mf} and Section~\ref{sec:mr}.

\textbf{Objective formulation.} While the objective term has different implementations for different applications, in general, we look for the optimal subset $\calP^* \subset \calP_0$: 

\begin{equation}
\calP^* = \argmin_{\calP \subset \calP_0} \calL(\calP),
\label{eq:obj_opt}
\end{equation}

where $\calL(\calP)$ is a general objective term:

\begin{equation}
\calL(\calP) = \Ls(\calP, I) + \Lreg(\calP) \> ,
\label{eq:obj_term}
\end{equation}

$\Ls(\calP, I)$ evaluates the error of selected proposals $\calP$ with respect to some input scene $I$. $\Lreg(\calP)$ evaluates error that is inversely proportional to the joint probability of proposals in $\calP$.

\section{MCTS with Refinement for Proposals Selection Games}
\label{sec:mcts_ref}

Monte Carlo Tree Search~(MCTS) is an algorithm used to efficiently search tree structures. To emphasize complexity of our experimental setting and the importance of efficient search algorithms, we visualize an example of a tree that our method builds in Figure~\ref{fig:mcts_tree}. We briefly describe the algorithm in this section and refer the interested reader to the survey in \cite{browne2012survey}. We also provide more details on our MCTS implementation in appendices. It has several advantages over other search techniques:

\begin{itemize}
    \item Because of its stochastic nature it is able to both exploit and explore the solution space more efficiently compared to other exploitation based algorithms, \eg hill-climbing.
    \item Evaluations are performed in leaf nodes of the tree. Hence, we can use very general objective functions that evaluate complete solutions, in comparison to unary and binary terms of graph-search algorithms.
    %
    %
    \item MCTS generalizes well to high-state-complexity problems.
\end{itemize}

\textbf{Scene Tree.} We follow~\cite{MCSS} and define a general tree structure. Each node in the tree corresponds to selecting a specific proposal from the pool and adding it to the set of selected proposals $\calP$. Sibling nodes in the tree are mutually incompatible, and a children node is compatible to all of its ancestors. One of the sibling nodes is always an \textit{empty} node which enables the algorithm to still find the optimal solution in case \textit{all} sibling nodes are false positives. Such tree structure ensures that following a path from root to any leaf node of the tree results in a valid solution respecting the rules of the proposals selection game.

\textbf{Scene Tree Traversal.} MCTS builds the tree online from a root. In a single iteration MCTS starts from the root node and traverses the known levels of the tree for which all nodes have been visited at least once. As each node contains its expected score based on previous iterations and the number of traversals in previous iterations, MCTS selects the nodes and descends the tree based on the Upper Confidence Bound~(UCB) term, that balances exploitation and exploration during the search. As soon as reaching an unknown node, \ie a node that has not been visited in previous iterations, MCTS enters the simulation phase. More exactly, it randomly selects proposals from the remaining pool, following the rules of the game, until the game ends. Then, we calculate the score, in our case $-\calL(\calP)$, for the selected proposals $\calP$ and update scores of nodes that were visited in the iteration. In practice, for each node we take maximum between the calculated score and the current expected score of the node.

\textbf{Refinement Phase.} We encounter situations in our applications where initially sub-optimal solutions may not fit the scene exactly, and without further refinement the expected score is relatively low and not reflecting the actual quality of the selected proposals. Therefore, after every score calculation we perform gradient descent based on objective term in Equation~\ref{eq:obj_term} and update proposals that were selected in this iteration.

\textbf{Final solution inference.} After $N=500$ MCTS iterations, we
traverse the tree selecting nodes with highest expected scores at each level. Finally, we optimize the selected proposals by minimizing the objective term. 
\section{Application 1: MonteFloor}
\label{sec:mf}

\newlength{\mrnicemethodwidth}
\setlength{\mrnicemethodwidth}{0.25\linewidth}
\begin{figure*}
    \centering
    {\footnotesize
    \begin{tabular}{cccccc}
        \includegraphics[height=\mrnicemethodwidth]{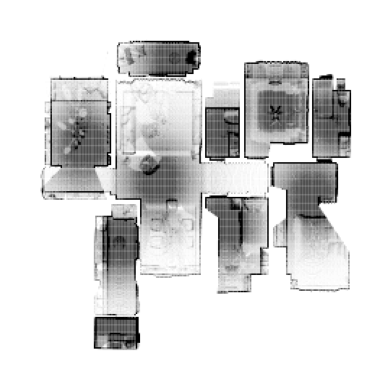} &
        \includegraphics[height=\mrnicemethodwidth]{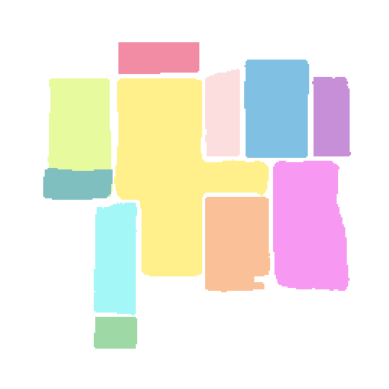} &
        \includegraphics[height=\mrnicemethodwidth]{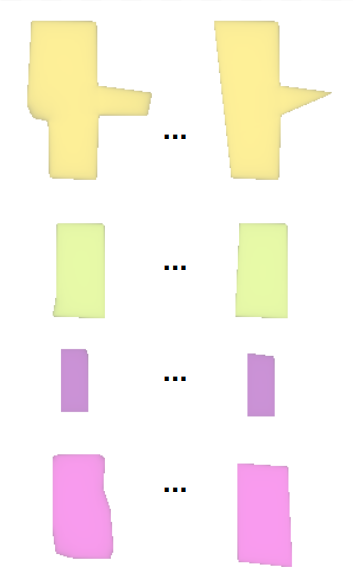} &
        \includegraphics[width=\mrnicemethodwidth]{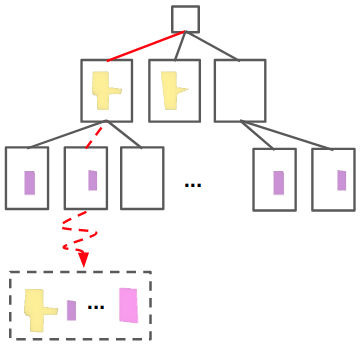} & 
        \includegraphics[height=\mrnicemethodwidth]{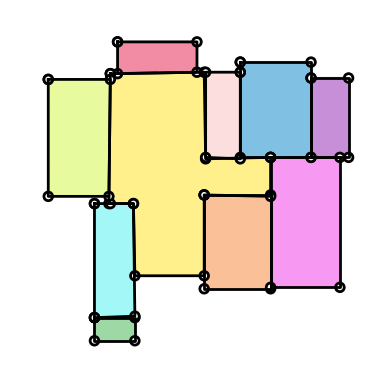} &
        \includegraphics[height=\mrnicemethodwidth]{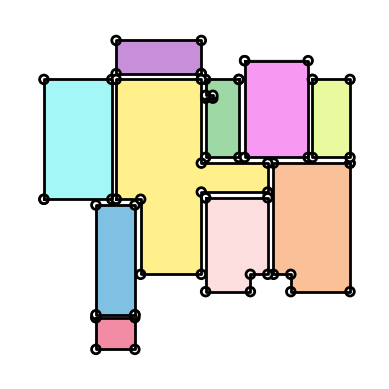} \\

        Density map & Detected & Some room proposals & Room selection with MCTS & Final result & Ground Truth\\[-0.05cm]
        & room segments &from polygonization  & + refinement & \\[-0.05cm]
        &&of the room segments&&\\[0.2cm]
    \end{tabular}
    }
    \caption{{\bf Overview of our MonteFloor method. } Given a 3D point cloud, we first create a density map of a floor. We then detect room segments using Mask-RCNN as in Floor-SP~\cite{chen2019floor}. Note the false positive at the bottom of the green segment on the left hand side. We polygonize each segment in different ways and obtain multiple room proposals from each room segment. We rely on MCTS and our objective function to select the correct room proposals, and our refinement step to adjust jointly the shapes of the room proposals to the input density map.}
    \label{fig:mf_method}
\end{figure*}

In this section we present an application of our approach to floor plan reconstruction that we refer to as MonteFloor. Given a point cloud of an indoor scene, we would like to accurately reconstruct extents of individual rooms. In top-view perspective of the scene we represent individual rooms as simple 2D polygons that define the 2D floor plan of the scene.

\textbf{Computing the density map representation.} As, from a top-view perspective, our setting is essentially reduced to a 2D vision problem, we first create a top-view density map $I$ from the input point cloud. We follow a similar procedure to the one in~\cite{chen2019floor}. First, from a given set of registered
 RGB-D panorama images, we generate a point cloud
of the scene. Then, in the top-view of the scene’s point cloud,
we project 3D points into a $256\times256$ image space,
such that the perspective is preserved and the complete scene is visible in the image. We define pixel's density value as the number of points that project to that pixel location normalized across the image to range $[0, 1]$.

We show an overview of our MonteFloor method in Figure~\ref{fig:mf_method}. We train Mask R-CNN~\cite{he2017mask} to predict rooms segments from input density maps. We polygonize predicted segments and obtain multiple polygonal approximations for each of the detected rooms. We call them the room proposals. Then, we use MCTS to find the subset of room proposals that is the best fit to the input density map. The selection is guided by a 'metric network' that is trained to estimate the quality of solution in respect to the input density map. As initial polygons only coarsely approximate the actual rooms outlines in the density map, we optimize polygonal proposals sampled during search iterations to improve overall fitness. In this context, we introduce a differentiable method for rendering general polygonal shapes.  

In following sections we explain how we generate room proposals, how we apply MCTS to select the optimal proposals, how we implement our objective term, and how we render polygonal room proposals during refinement.  

\subsection{Generating Room Proposals}

From a density map of the scene, we infer individual room segments using a Mask R-CNN~\cite{he2017mask}, which we trained on the training scenes of the Structured3D dataset~\cite{Structured3D}. Even though we obtain high quality room segments, we still need to determine the correct polygonal shapes and remove false positive detections. Figure~\ref{fig:mf_method} shows an example of room segments and the room proposals we generate from them. We detail this process below.

Sometimes, a room is detected as two segments that partially overlap. We thus merge two segments that overlap significantly (more than $5\%$ in practice) into an additional room segment, while keeping the two original segments.

In practice, the shapes of the true positive segments provided by Mask-RCNN do not correspond to the exact shapes of the rooms, as they are typically too smooth. We thus  polygonize the room segments to generate the room proposals. For this, we apply the Douglas-Peucker polygonization algorithm~\cite{douglas1973algorithms} to the contours of the room segments. This algorithm depends on a parameter $\epsilon$ that controls the simplification of the contour. More exactly, this parameter is the maximum distance between the original curve and its approximation. As the exact complexity of the room shape is unknown at this stage, we generate multiple proposals from each segment by using different values for $\epsilon$. In practice, we take $\epsilon = d \cdot L$, where $d$ takes different values in a predefined set $\calD$ and $L$ is the perimeter of the segment, with $\calD = \{0.04, 0.02, 0.01\}$. Sometimes, 2 different $\epsilon$ result in the same number of vertices, and we keep only one of the two polygons.

Even after polygonization, the shapes of the true positive room proposals may not correspond yet to the actual room shapes. To adjust their shapes, we will refine them through our objective function. Without this refinement it is possible that the value of the objective function is relatively low and not
reflecting the actual quality of the selected proposals well.
The refinement step adjusts the locations and shapes of the
room proposals to obtain a more accurate solution. We describe the room proposal selection game in the next subsection, and our implementation of the objective term afterwards.

\subsection{Room Proposals Selection Game}

\newlength{\mfprogresstwidth}
\setlength{\mfprogresstwidth}{0.10\linewidth}
\begin{figure*}
    \centering 
    \scalebox{0.85}{
    \begin{tabular}{ccc}
         \includegraphics[height=0.25\linewidth]{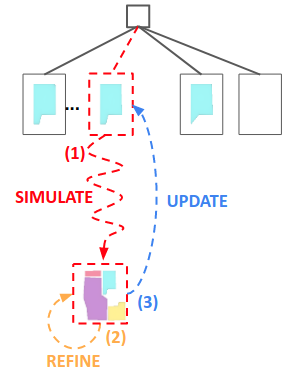} &  
         \includegraphics[height=0.25\linewidth]{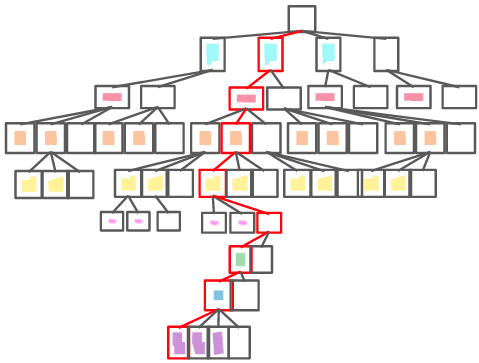} &
         \begin{tabular}[b]{ccccc}
         \multicolumn{5}{c}{\includegraphics[width=0.4\linewidth, trim=15 5 5 5, clip]{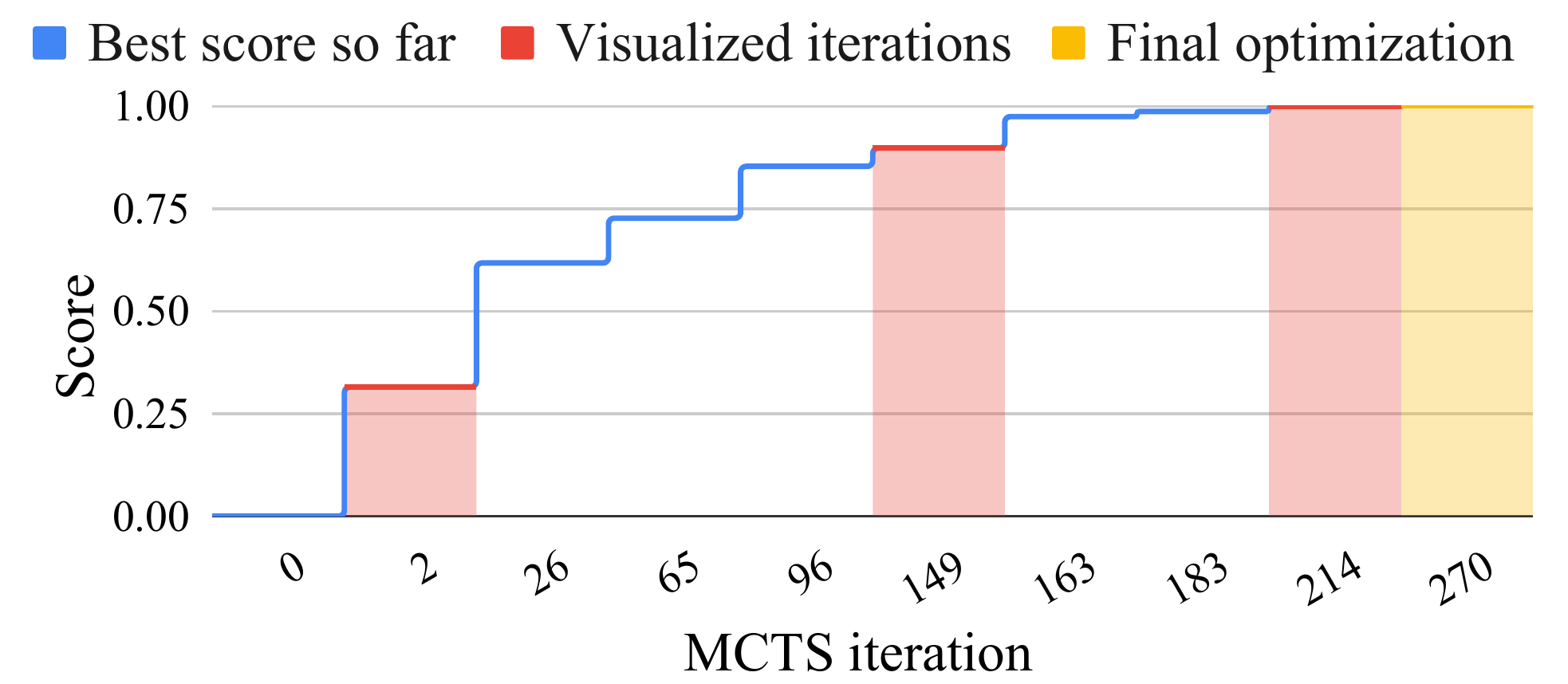}} \\
         \includegraphics[width=\mfprogresstwidth]{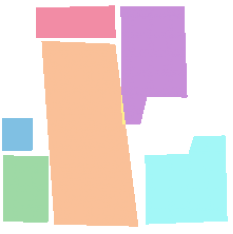} & 
         \includegraphics[width=\mfprogresstwidth]{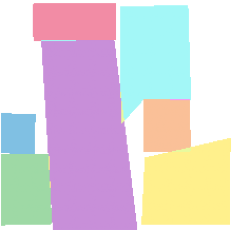} & 
         \includegraphics[width=\mfprogresstwidth]{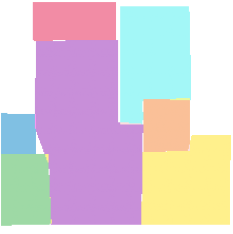} &
          \includegraphics[width=\mfprogresstwidth]{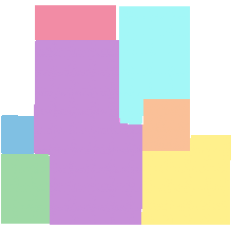} & 
         \includegraphics[width=\mfprogresstwidth]{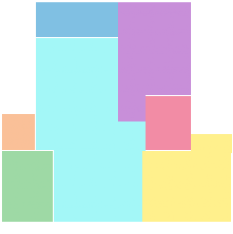} \\
         iter. 2 & iter. 149 & iter. 214 & final & g.t. \\
        \end{tabular} \\
         (a) & (b) & (c) \\
    \end{tabular}}
    \caption{{\bf Building the floor plan tree with MCTS}. (a) In our case, one node corresponds to the selection of a room proposal, or to skipping all the room proposals generated from a room segment. When a node is visited for the first time, MCTS runs a 'simulation' step. This step explores randomly the rest of the tree until reaching a leaf, in our case when there is no room proposal to consider any more. When reaching a leaf, we perform our 'refinement step', which optimizes the objective function over the room proposals in the path from the root node to the leaf. The value of the objective function is used to update the expected score for all the nodes in the path. (b) MCTS builds and explores only a portion of the tree. In contrast with other tree search algorithms, the pruning of MCTS is based only on the evaluations of leaves, which means that we can select a set of proposals based on how well they explain the density map together. (c) After few iterations, our MonteFloor method focuses at, and at the same time optimizes, solutions with promising expected scores. This enables us to quickly reconstruct an accurate floor plan of the scene, in about $60$ seconds for the scene used in this illustration.}
    \label{fig:tree}
\end{figure*}

As shown in Figure~\ref{fig:tree}, in our room proposal selection game, a move consists of selecting exactly one of the room proposals generated by polygonizing one of the room segments. By rephrasing this in terms of a general proposal selection game, we define proposals generated from a single room segment to be mutually incompatible. For each room segment, there is an additional move that consists of not selecting any of the room proposals from this segment to enable removal of false positive room detections. As in the general version, the game ends after there are no more available proposals in the pool. 

\textbf{Scene Tree complexity.} The root node has up to $|\calD|+1$ children, corresponding to the  selection of one of the $|\calD|$ room proposals issued from the first room segment and the absence of selection from this room segment. The number of nodes of the full tree is at most $(|\calD|+ 1)^k$ where $k$ is the number of room segments, and, as $k$ increases, it quickly becomes infeasible to traverse all paths in the tree. Fortunately, MCTS will grow the tree only as needed while exploring it and it avoids an exhaustive evaluation. 

\textbf{Final solution post-processing.}
For some rare polygons, the vertices are less than $5$ pixels apart from each other. We merge the corresponding vertices to obtain the final solution.

\subsection{Implementing the Scene Loss}

In our MonteFloor application, the scene loss $\Ls(I,\calP)$ evaluates quality of the subset of proposals $\calP$ with respect to the density map $I$:

\begin{equation}
\Ls(I, \calP) = - \lambda_f f(I, F(P)) \> ,
\label{eq:mf_obj_fun}
\end{equation}

where $f(I, F(\calP))$ is our metric network, applied to the input density map and the floor plan of the room proposals $\calP$, weighted by $\lambda_f$. 


\begin{figure}
    \centering
    \includegraphics[width=0.8\linewidth]{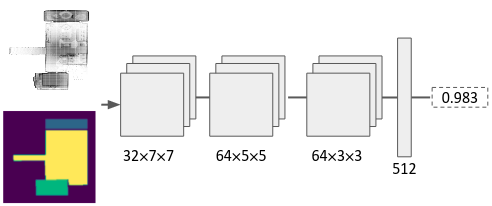}
    \caption{{\bf Our metric network.} This network takes a density map and a representation of the floor, colorized for visualized purposes, as input and outputs a score that measures how well the floor map fits the input density map. We train it to predict the Intersection-over-Union between the estimated floor plan and the ground truth.
    }
    \label{fig:scorenet}
\end{figure}

\subsection{Metric Network}
\label{sec:mf_metric_network}

Our metric network $f(I, F(\calP))$ evaluates how well a set $\calP$ of selected room proposals fits the input density map $I$. As shown in Figure~\ref{fig:scorenet}, this network has a simple architecture and takes two inputs: The first input $I$ is the density map. The second input $F(\calP)$ is an image of the room proposals, which we render  using their indices as pixel values:
\begin{equation}
F(\calP) = \sum_i i R(P_i) \> ,
\end{equation}
where $R(P_i)$ is a binary image of $P_i$ for which the pixels inside $P_i$ are set to 1 and the others to 0. 

$f$ outputs only a single value, which should reflect the fitness between the room proposals and the density map. We train it to predict the Intersection-over-Union~(IOU). 

\textbf{Training the metric network.} The metric network is trained on the Structured3D dataset~\cite{Structured3D} with input density maps and floor plans of size $256 \times 256$. During training, we generate input floor plans directly from the ground truth annotations to simulate a large variety of possible settings. More exactly, with probability of $30 \%$, we select the ground truth floor plan as input. Otherwise, each individual room is added with probability of $50 \%$. A single room is randomly rotated with $50 \%$ chance by either $90^\circ, 180^\circ$, or $270^\circ$, and randomly translated by $[-50, 50]$ pixels with probability of $10 \%$. With $1 \%$ probability, a vertex inside a room polygon is translated in range $[-10, 10]$ pixels. Labels for individual rooms are shuffled to make sure the network does not overfit to some specific label ordering. Similarly to input floor plans, we augment the dataset by rotating and translating the input density map and the corresponding ground truth floor plan. We match the generated room shapes with the ground truth shapes, as in Section 4.1 of the main paper, and calculate the Intersection-Over-Union (IOU) between the matches to obtain the final ground truth score.

The network is then trained by minimizing the Root-Mean-Square-Error (RMSE) between predicted and ground truth scores. We use the Adam optimizer~\cite{kingma2014adam} with learning rate set to $10^{-3}$.


\subsection{Implementing the Regularization Loss}

\newcommand{\ot}{\frac{1}{2}}
\newcommand{\of}{\frac{1}{4}}
\newcommand{\ps}{\frac{\pi}{6}}
\newcommand{\pt}{\frac{\pi}{2}}
\newcommand{\fpt}{\frac{5\pi}{12}}
\newcommand{\fps}{\frac{5\pi}{6}}
\newcommand{\sps}{\frac{7\pi}{6}}

\newcommand{\ang}{\text{ang}}
\newcommand{\reg}{\text{reg}}
\newcommand{\glob}{\text{glob}}
\newcommand{\TV}{\text{TV}}
\newcommand{\MSE}{\text{MSE}}

We implement $\Lreg(\calP)$ to regularize polygonal room proposals and decompose it to: 
\begin{equation}
    \Lreg(\calP) = \lambda_\ang \calL_\ang(\calP) + \lambda_\glob \calL_\glob(\calP) + \lambda_0 \calL_0(\calP) \> ,
\end{equation}
where $\lambda_\ang$, $\lambda_\glob$, and $\lambda_0$ weight the three terms.

$\calL_\ang(\calP)$ regularizes the angles of the room proposals in $\calP$:
\begin{equation}
    \calL_\ang(\calP) = -\frac{1}{|\calP|}\sum_{P_i \in \calP} \frac{1}{|P_i|} \sum_{(u,v,w) \in P_i} \log p(\widehat{(u,v,w)} ) \> ,
\end{equation}
where $|P_i|$ denotes the number of vertices in polygon $P_i$, $(u,v,w)$ denote any three consecutive vertices of polygon $P_i$, and $\widehat{(u,v,w)}$ their angle at vertex $v$. $p(\alpha)$ is a prior distribution we assume over the room angles. As shown in Figure~\ref{fig:mf_reg}, we use a mixture of Gaussian distributions over their cosine and uniform distributions. It discourages flat angles ($0\degree$ and $180\degree$), encourages right angles (90$\degree$ and 270$\degree$), and angles between $\pi/6$ and $5\pi/6$ and between $7\pi/6$ and $-\pi/6$ follow a uniform distribution. More formally, we take $p(\alpha) = $

{\normalsize
\begin{equation}
\frac{1}{Z}
\left\{
\begin{array}{ll}
G(\cos{\alpha} \;|\; \cos{\ps}, \sigma_1) & \text{if } \alpha \in ]-\ps;\ps] \>,\\ [0.1cm]
\eta + G(\cos{\alpha} \;|\; \cos{\pt}, \sigma_2) & \text{if } \alpha \in ]\ps;\fps]\>, \\[0.1cm]
G(\cos{\alpha} \;|\; \cos{\fps}, \sigma_1) & \text{if } \alpha \in ]\fps;\sps]\>, \text{ and} \\[0.1cm]
\eta + G(\cos{\alpha} \;|\; \cos{\-\pt}, \sigma_2) & \text{if } \alpha \in ]\sps;-\ps] \> ,\\[0.1cm]
\end{array}
\right.
\end{equation}
}
where $G$ denotes Gaussian distribution, $\eta$ is the constant $G(\cos{\ps} \;|\; \cos{\ps}, \sigma_1)$, and $Z$ is a normalization factor. In practice, we use $\sigma_1=0.1$ and $\sigma_2=0.08$.

$\calL_\glob(\calP)$ encourages room proposals to be in contact without overlapping. It can be seen in Figure~\ref{fig:mf_reg} that the Total Variation~(the sum of the absolute values of the gradients) of an image of the proposals is a good criterion:
\begin{equation}
    \calL_\glob = \TV(F_1(\calP)) \> ,
\end{equation}
where $\TV$ denotes the total variation and $F_1(\calP)$ is an image of the proposals computed as 
\begin{equation}
F_1(\calP) = \sum_i R(P_i) \> .
\end{equation}
Figure~\ref{fig:mf_reg} shows that this loss penalizes overlaps and pushes proposals toward each other, and by doing so, enforces similar orientations between the walls of neighbouring rooms.

\newlength{\tvwidth}
\setlength{\tvwidth}{0.2\linewidth}
\begin{figure}
    \centering
    {\footnotesize
    \begin{tabular}{@{}c@{}c@{}}
    \adjustbox{valign=c}{\includegraphics[width=0.35\linewidth,trim=30 5 50 10, clip]{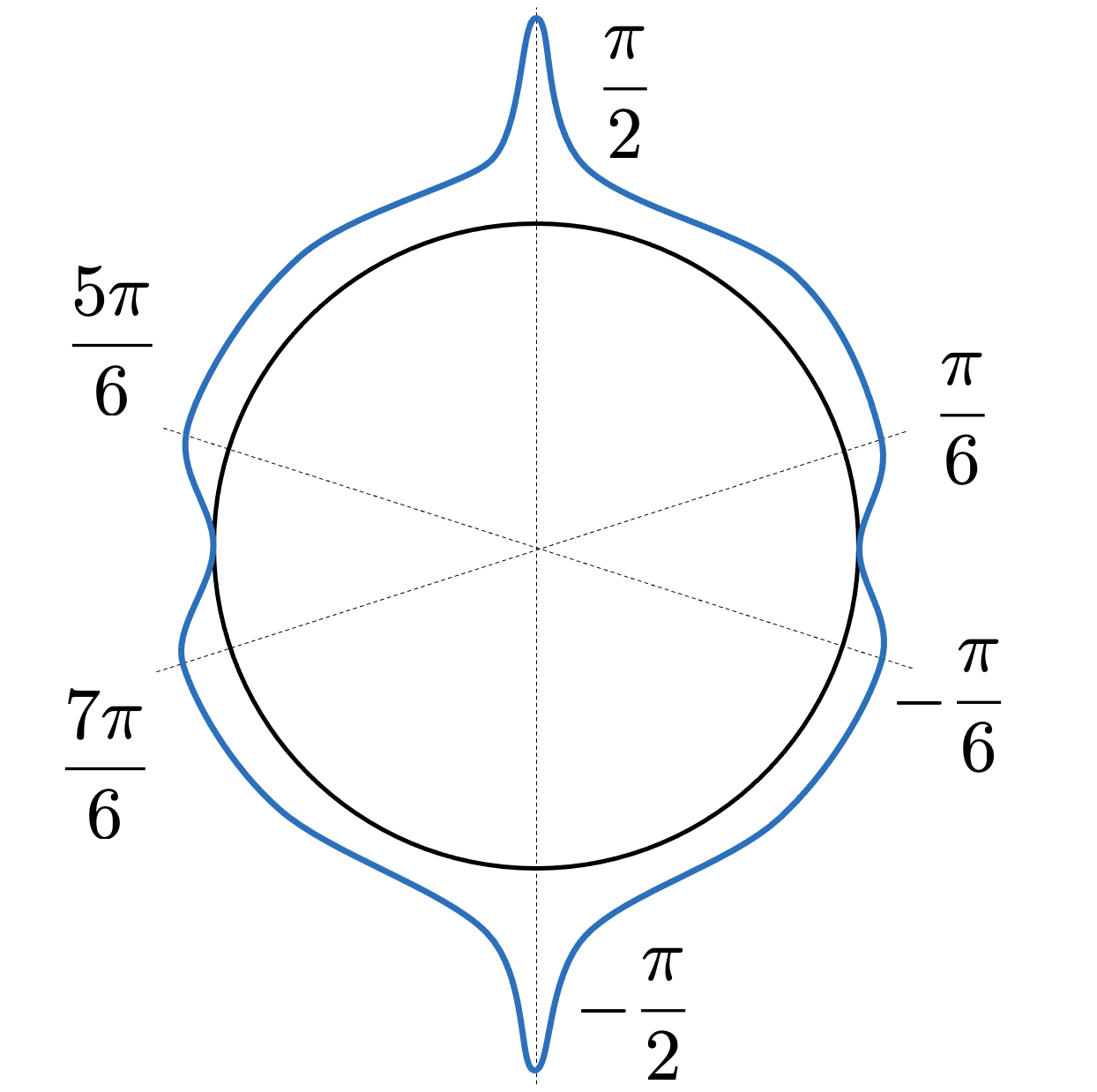}} &
    \begin{minipage}{0.6\linewidth}
        \begin{tabular}{@{}c@{}c@{}c@{}}
             \includegraphics[width=\tvwidth]{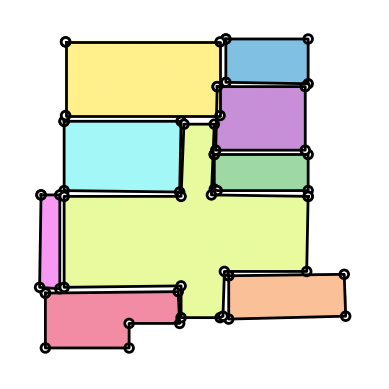}  &
             \includegraphics[width=\tvwidth]{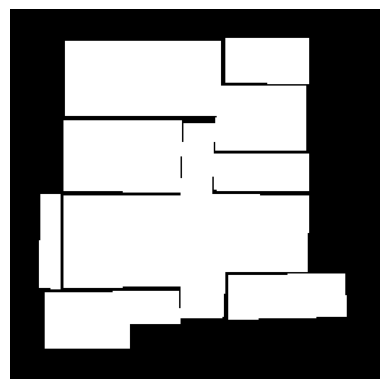}  &
             \includegraphics[width=\tvwidth]{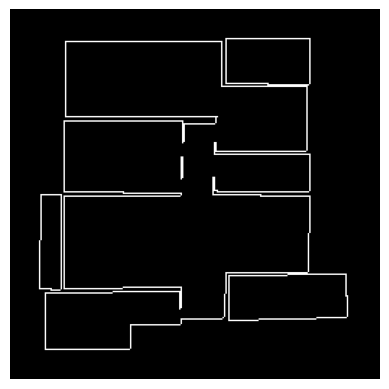}  \\
             
             \includegraphics[width=\tvwidth]{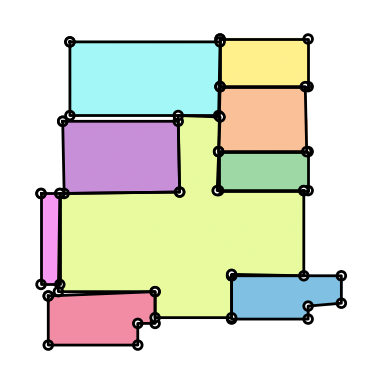}  &
             \includegraphics[width=\tvwidth]{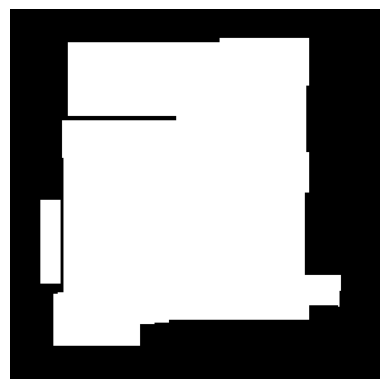}  &
             \includegraphics[width=\tvwidth]{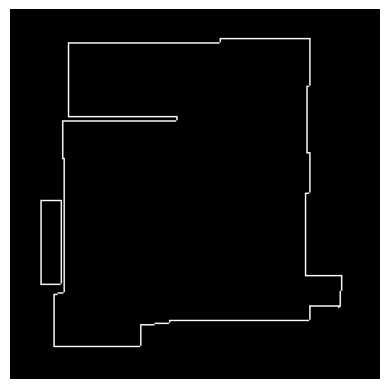}  \\

             $P$ & $F_1(\calP)$ & per pixel \\
                 &          & $\TV(F_1(\calP))$ \\
        \end{tabular}
    \end{minipage} \\
    (a) & (b)
    \end{tabular}
    }
    \caption{{\bf Visualization of the regularization losses.} (a) Prior distribution $p(\alpha)$ on angles discourages flat angles and encourages right angles, but other angles can still be accepted. (b) $\calL_\glob$ is based on total variation. {\bf Top:} When the room proposals in $\calP$ are not in contact or overlap, the Total Variation $\TV(F_1(\calP))$ of their image $F_1(\calP)$ is large. {\bf Bottom: } When the room proposals fit together, the Total Variation $\TV(F_1(\calP))$ is much lower.}
    \label{fig:mf_reg}
\end{figure}

$\calL_0$ is used to prevent the proposals to drift from their initial locations during optimization. We take:
\begin{equation}
    \calL_0(\calP) = \frac{1}{|\calP|}\sum_{P_i \in \calP} \MSE(R(P_i), M_i) \> ,
\end{equation}
where $M_i$ is the binary image of the segment that generated proposal $P_i$, and $\MSE(\cdot)$ compares this image with the binary image $R(P_i)$ of the proposal.


\subsection{Differentiable Polygon Rendering}

As explained earlier, when MCTS reaches a leaf, we perform several optimization steps of the objective function in Eq.~\eqref{eq:mf_obj_fun} before computing its value and using it as a score for MCTS. In practice, we use the Adam optimizer~\cite{kingma2014adam}  for this task.

To optimize $\calL(\calP)$, we need to make it differentiable. The only part of it that is not trivially differentiable is the binary image creation $R(P_i)$ of a room proposal $P_i$, where $P_i$ is represented as a polygon. Differentiable rendering has already been developed~\cite{OpenDR}, however, available implementations are designed for rendering meshes of 3D triangles. Instead of tweaking these implementations to make them work on 2D polygons, we developed a much simpler approach by making the winding number algorithm differentiable. The original winding number algorithm checks whether a pixel location $m$ is inside a polygon $P_i$ by computing:
\begin{align}
    W(m, P_i) & = \frac{1}{2\pi}\sum_{(u,v) \in P_i} \text{sign}(\det(um, vm)) \widehat{(umv)} \> ,
\end{align}
where $(u,v)$ are any 2 consecutive vertices of $P_i$ and $\det(\cdot)$ is the determinant of vectors $um$ and $vm$. The $\text{sign}(\cdot)$ term is equal to 1 if angle $\widehat{(umv)}$ is between $]0;\pi]$ and to -1 if it is between $]\pi;2\pi[$, and $0$ otherwise. Hence, for a valid non-intersecting, closed, and counter-clockwise oriented polygon, $W(m, P_i) \in \{0,1\}$,  is a step function with value $1$ if $m$ is inside of $P_i$ and $0$ otherwise. 

To make it differentiable, we use the following expression instead:
\begin{align}
    W(m, P_i) & = \frac{1}{2\pi}\sum_{(u,v) \in P_i}
    \frac{c \cdot \det(um, vm)}{1 + |c \cdot \det(um, vm)|} \widehat{(umv)} \> .
\end{align}

The fraction term implements a soft form of the sign function that measures orientation of the triangle $(umv)$ with $c=1000$ to approximate the step form of the sign function in a differentiable way. To make rendering more efficient, we calculate the winding values only for pixel locations $m$ that are inside the bounding box detected by Mask R-CNN for the corresponding room segment.


\section{MonteFloor Evaluation}

In this section, we evaluate our MonteFloor method by comparing it to Floor-SP~\cite{chen2019floor}, the current state-of-the-art in floor plan reconstruction, on two datasets. We also provide an ablation study to show the importance of the refinement step for our method. 

\textbf{Choice of Hyper-Parameters for the Objective Term.} During refinement step, we set $\lambda_\ang=0.01, \lambda_0=0.01, \lambda_\glob=0.2, \lambda_f=0.1$ and normalize $p(\alpha)$ to $[0, 1]$. We have found empirically that this set of hyper-parameters balances the influence of the corresponding loss terms. Setting $\lambda_\glob=0$ and $\lambda_f=1$ in the score calculation step increases convergence speed.

For the test set from \cite{chen2019floor}, our metric network is less stable as it was not trained on this dataset. Hence, we have observed that setting $\lambda_\ang=0.1, \lambda_0=0.1, \lambda_\glob=0.2, \lambda_f=0.05$, during refinement step, improves performance. During the score calculation step, we still set $\lambda_\glob=0$ and $\lambda_f=1$.

\subsection{Metrics} 

To evaluate the quality of recovered floor plans, we first match the recovered rooms to the ground truth rooms. More exactly, starting with the largest ground truth room, we find the matching recovered room with the largest Intersection-Over-Union~(IOU) value. As we believe that the metrics used in~\cite{chen2019floor} are too permissive for really evaluating the quality of the compared approach, we made them more strict for quantitative evaluation:
\begin{enumerate}
    \item \textbf{Room metric.} This metric is the same as in \cite{chen2019floor}. A room polygon is considered to be successfully recovered if it is not overlapping other rooms and if it is matched with a ground truth room. We allow one pixel overlap between rooms and hence we do not penalize room polygons that are touching with this metric.
    
    \item \textbf{Corner metric.} A corner is considered to be successfully recovered if its corresponding room polygon is successfully recovered and it is the closest corner to any of the corners in the matching ground truth room polygon, within a distance of 10 pixels. This metric is inspired by the original metric from~\cite{chen2019floor} that did not consider if the corner actually belongs to the correct polygon.
    
    \item \textbf{Corner angle metric.} An angle of a room polygon is considered to be successfully recovered if its corresponding corner is successfully recovered and if the absolute difference to the corresponding ground truth angle is less than $5^\circ$.
\end{enumerate}

\begin{table}
    \centering
    \scalebox{.85}
	     {
    \begin{tabular}{@{}l|c c c c c c | c c @{}}
    & \multicolumn{2}{c}{Room} & \multicolumn{2}{c}{Corner} & \multicolumn{2}{c|}{Angle} & \multicolumn{2}{c}{MA} \\
    & Prec & Rec &  Prec & Rec & Prec & Rec & Prec & Rec \\
    \hline
    \multicolumn{1}{l | }{{\bf Structured3D}}&\multicolumn{6}{l|}{}\\
    DP ($\epsilon = 0.01$) & 0.93  & 0.94  & 0.74  & 0.79  & 0.49  & 0.52  & 0.72  & 0.75  \\
    Floor-SP~\cite{chen2019floor}     & 0.89  & 0.88  & 0.81  & 0.73  & 0.80  & 0.72  & 0.83  & 0.78  \\

    MonteFloor (ours)  & \textbf{0.96}  & \textbf{0.94}  & \textbf{0.89}  & \textbf{0.77}  & \textbf{0.86}  & \textbf{0.75}  & \textbf{0.90}  & \textbf{0.82}  \\
    
    \hdashline
    HEAT~\cite{chen2021heat} {\bf *} & 0.97 & 0.94 & 0.82 & 0.83 & 0.78 &  0.79 & 0.86 & 0.85 \\
    
    \hline
    \multicolumn{1}{l|}{{\bf \cite{chen2019floor} test set}}&\multicolumn{6}{l|}{}\\

    Floor-SP~\cite{chen2019floor}  & 0.85  & 0.83  & 0.72  & 0.58  & 0.65  & 0.52  & 0.74  & 0.64  \\

    MonteFloor (ours)  & \textbf{0.88}  & \textbf{0.85}  & \textbf{0.78}  & \textbf{0.63}  & \textbf{0.68}  & \textbf{0.54}  & \textbf{0.78}  & \textbf{0.67}  \\
    
    \end{tabular}}
    \caption{Quantitative evaluation of MonteFloor on Structured3D~\cite{Structured3D} and the test set from~\cite{chen2019floor}. MA is the average of the three metrics~(Room, Corner, and Angle). We compare our approach to a simple  Douglas-Peucker polygonization of the room segments obtained by Mask-RCNN~(DP) and to Floor-SP~\cite{chen2019floor}. Our approach slightly outperforms the other methods, even though we could not train our metric network on the training set of \cite{chen2019floor}. \textbf{*} HEAT~\cite{chen2021heat} is a very recent method that performs comparably to our method.}
    \vspace{-1cm}
    \label{tab:mf_quant}
\end{table}

\subsection{Evaluation and Comparison with Floor-SP}

\noindent \textbf{Structured3D.} We perform a first evaluation on the Structured3D dataset~\cite{Structured3D} that contains floor plan annotations for $3500$ scenes: $3000$ training scenes, $250$ validation scenes, and $250$ test scenes. To mimic the standard scene reconstruction pipeline, we project the registered RGB-D panorama images to obtain the point cloud of the scene. We process the reconstructions to obtain the training data for both Mask R-CNN and metric networks. For a fair comparison, we retrained the network used by Floor-SP for predicting the corner -and edge- likelihood maps on the training set generated from the Structured3D dataset and we replaced their Mask R-CNN network by ours, which was also trained on Structured3D.

\noindent \textbf{Floor-SP test set.} Unfortunately, the authors of Floor-SP~\cite{chen2019floor} could not publish the training scenes for their Floor-SP dataset, but we could evaluate our approach on the $100$ publicly available test scenes, which include a large variety of floor plan configurations. We use the Mask R-CNN network pretrained on their training set as it was made available by the authors. However, since we could not train our metric network on the Floor-SP training set, we had to use the one trained only on Structured3D. Hence, the Floor-SP method has an advantage on this dataset.

\newlength{\mfniceresultwidth}
\setlength{\mfniceresultwidth}{0.18\linewidth}
\newcommand{\mfniceresult}[1]{
\includegraphics[width=\mfniceresultwidth]{figures/mf_s3d_qual/#1_density.png} & 
\includegraphics[width=\mfniceresultwidth]{figures/mf_s3d_qual/#1_fsp_edit.png} &
\includegraphics[width=\mfniceresultwidth]{figures/mf_s3d_qual/#1_ours.png} &
\includegraphics[width=\mfniceresultwidth]{figures/mf_s3d_qual/#1_gt.png} \\}

\newlength{\mfniceresultwidthfsp}
\setlength{\mfniceresultwidthfsp}{0.18\linewidth}
\newcommand{\mfniceresultfsp}[1]{
\includegraphics[width=\mfniceresultwidthfsp]{figures/mf_fsp_qual/#1_density.png} & 
\includegraphics[width=\mfniceresultwidthfsp]{figures/mf_fsp_qual/#1_fsp.png} &
\includegraphics[width=\mfniceresultwidthfsp]{figures/mf_fsp_qual/#1_ours.png} &
\includegraphics[width=\mfniceresultwidthfsp]{figures/mf_fsp_qual/#1_gt.png} \\}

\begin{figure}
    \centering
    \begin{tabular}{cccc}
    \mfniceresult{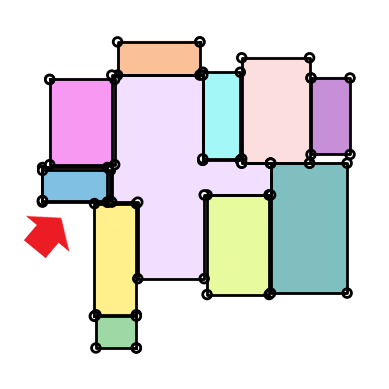}
    \mfniceresult{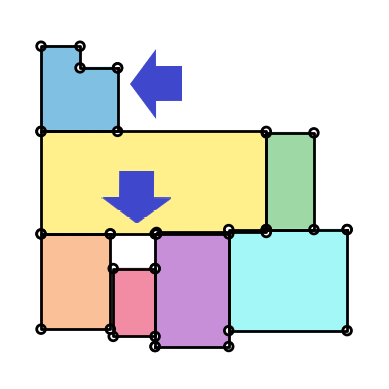}
    \mfniceresult{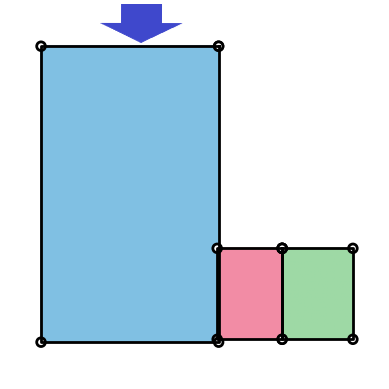}
    Input & Floor-SP & Ours & Ground Truth \\
    \end{tabular}
    \caption{\textbf{Qualitative results on the Structured3D dataset~\cite{Structured3D}, best seen in color.} Red arrow: In contrast to Floor-SP, our approach deals well with false positive detections. Blue arrows: Compared to Floor-SP, we are able to model a larger variety of room shapes.
    }
    \label{fig:mf_qual_sfp}
\end{figure}

\begin{figure}
    \centering
    \begin{tabular}{cccc}
    \mfniceresultfsp{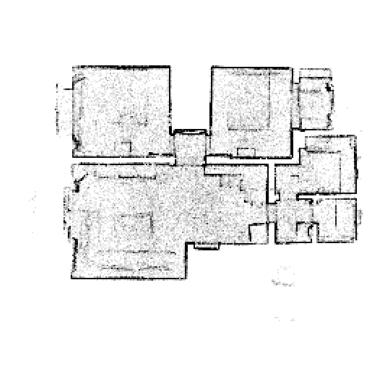}
    \mfniceresultfsp{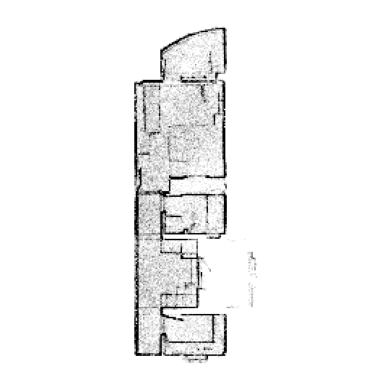}
    \mfniceresultfsp{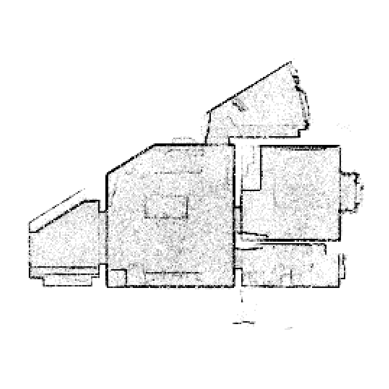}

    Input & Floor-SP & Ours & Ground Truth \\
    \end{tabular}
    \caption{\textbf{Qualitative results on the test set from~\cite{chen2019floor}}. Even though our metric network was not trained on the Floor-SP training set, our method still performs slightly better than Floor-SP on the Floor-SP test set. Red arrow: We remove false positive room detections. Green arrow: The purple room in the ground truth appears to be an annotation error. Blue arrows: Our reconstructions are sometimes more consistent with the input than the manually annotated rooms.}
    \label{fig:mf_qual}
\end{figure}

Table~\ref{tab:mf_quant} shows the quantitative results on both datasets. To better demonstrate the benefits of our approach, we also show the results of simple room detection by polygonization of room masks detected by Mask R-CNN with the Douglas-Peucker~(DP) approach that we used to initialize room proposals. DP obtains very high performance for the room metric, indicating that Mask R-CNN outputs masks of good quality most of the time. However, the angle metric clearly demonstrates that these polygons very often do not look anything like the actual room shapes. 

For Floor-SP, there is a drop in the room metric in comparison to the Douglas-Peucker method. This is related to the containment constraint satisfaction in the Floor-SP approach that forces the retrieved polygons to contain the segmentation mask completely. If this constraint cannot be enforced, the reconstruction will also fail. However, more importantly, the angle metric clearly demonstrates that their results are still superior to the ones obtained by DP.

Our approach outperforms both baseline methods by a large margin as we maintain very high performance on all metrics. This is true even for the Floor-SP test set, even though we could not retrain our metric network on the corresponding training set.

We improve performance on the room metric in comparison to Douglas-Peucker method as our refinement step adjusts the shapes of the room proposals that may initially overlap, and the selection by MCTS removes false positives. In contrast to Floor-SP, our approach benefits from optimizing directly on polygon shapes that enables us to avoid both the mask containment and angle discretization constraints.

For completeness, we include very recent results from the HEAT method~\cite{chen2021heat} on the Structured3D dataset that reconstructs floor plans as a planar graph and performs comparably to our method. It would be of interest to combine these insights in future work as HEAT is very fast during inference. 

In addition, we compared the execution time of the two methods on the same machine. On the Structured3D dataset, the average computation time for Floor-SP is $785 \pm 549$ seconds. In contrast, the average computation time for our MonteFloor method is $71 \pm 40$ seconds, and $12 \pm 8$ seconds when skipping the refinement step. We made similar observations on the Floor-SP dataset.

\textbf{Qualitative results. } Figures~\ref{fig:mf_qual_sfp} and \ref{fig:mf_qual} show some qualitative results and  demonstrate that our approach is able to remove false positive detections and retrieve highly accurate polygonal reconstructions of floor plans.

\textbf{Ablation Study.} We performed an ablation study to evaluate the effectiveness of each individual term of our refinement procedure. As shown in Table~\ref{tab:mf_ablation}, all of our regularization terms help to retrieve room polygons of better locations and shapes. The main ablation shows that the metric network has also a crucial role in our approach. Without the metric network, the objective function does not enforce consistency with the input scene. Then choosing a single correct room in a large scene maximizes precision as there are indeed no false positives, but minimizes recall. 

\newcommand{\without}{w\textbackslash o}

\begin{table}
    \centering
    \scalebox{0.9}
	     {
    \begin{tabular}{@{}l|c c c c c c | c c @{}}
    & \multicolumn{2}{c}{Room} & \multicolumn{2}{c}{Corner} & \multicolumn{2}{c}{Angle} & \multicolumn{2}{c}{MA} \\
    
    & Prec & Rec &  Prec & Rec & Prec & Rec & Prec & Rec \\
    \hline
    no refin. step   & 0.95  & 0.93  & 0.86  & 0.76  & 0.65  & 0.57  & 0.82  & 0.75  \\

    w/o $\calL_\ang$     & 0.96  & 0.94  & 0.86  & 0.75  & 0.73  & 0.68  & 0.85  & 0.79  \\

    w/o $\calL_\glob$   & 0.85  & 0.84  & 0.78  & 0.69  & 0.74  & 0.66  & 0.79  & 0.73  \\

    w/o $\calL_0$   & 0.92  & 0.92  & 0.87  & 0.76  & 0.84  & 0.72  & 0.88  & 0.80  \\
    
    w/o $f(.)$  & 0.94  & 0.22  & 0.89  & 0.15  & 0.87  & 0.15  & 0.90  & 0.17  \\

    complete & 0.96  & 0.94  & 0.89  & 0.77  & 0.86  & 0.75  & 0.90  & 0.82  \\

    \end{tabular}}
    \caption{\textbf{Ablation study.} Removing $\calL_\ang$ has a large influence on the angle metric;  Removing $\calL_\glob$ has a large influence on the locations of the corners; Removing $\calL_0$ may result in drift. Our metric network $f(\cdot)$ is crucial for the selection step of MCTS as the other terms are not a suitable scoring function for the floor plan generation task.
    }  
    \label{tab:mf_ablation}
    \vspace{-0.5cm}
\end{table}

\section{Application 2: MonteRoom}
\label{sec:mr}

In this section we present an extension of our approach for the task of general 3D room layout estimation. More precisely, given an input color image we would like to precisely reconstruct extents of structural components, such as walls, floors, and ceilings, and their connectivities.

We show an overview of our MonteRoom method in Figure~\ref{fig:method_mr}. In following sections we explain how we formulate the 3D room layout estimation task, how we define and how we generate the initial set of proposals, and in the end, how we define our objective functions. Our scheme for jointly performing discrete optimization, using MCTS, and continuous optimization, using gradient descent, remains unchanged, further demonstrating the generalization aspect of our approach to other problems in 3D scene understanding.

\newlength{\mroverviewwidth}
\setlength{\mroverviewwidth}{0.18\textwidth}
\newlength{\mroverviewwidthhalf}
\setlength{\mroverviewwidthhalf}{0.08\textwidth}
\begin{figure*}
    \centering
    {\footnotesize
    \addtolength{\tabcolsep}{-0.1cm}
    \scalebox{0.95}{
    \begin{tabular}{cccccc}
    \multirow{2}{*}[1cm]{\includegraphics[width=\mroverviewwidth]{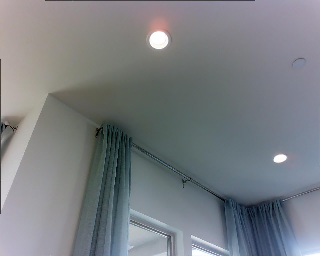}} & 
      \multirow{2}{*}[1cm]{\includegraphics[width=\mroverviewwidth]{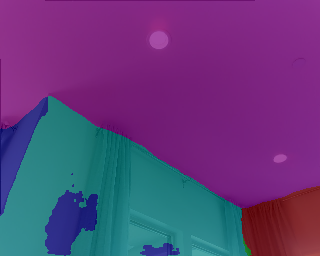}} & 

    \includegraphics[width=\mroverviewwidthhalf]{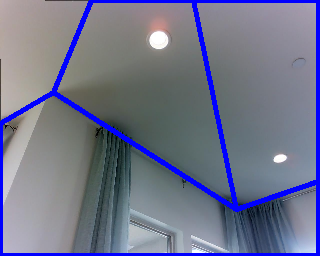} & \hspace{-0.3cm}
    \includegraphics[width=\mroverviewwidthhalf]{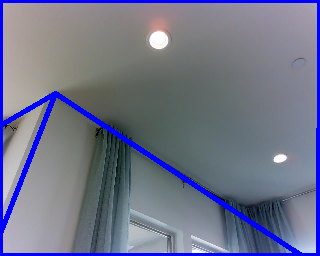} &
    \multirow{2}{*}[1cm]{\includegraphics[width=\mroverviewwidth]{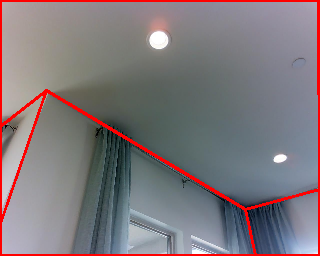}} &

    \multirow{2}{*}[1cm]{\includegraphics[width=0.23\textwidth]{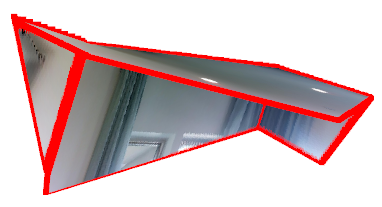}} \\
    
    & &
    
    \includegraphics[width=\mroverviewwidthhalf]{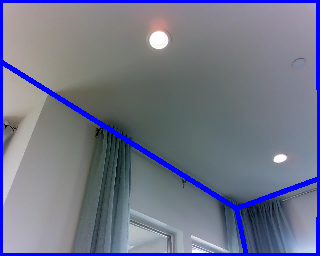} & \hspace{-0.2cm}
    \includegraphics[width=\mroverviewwidthhalf]{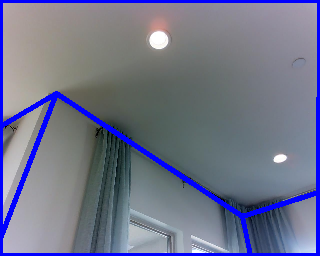} \vspace{0.1cm} \\
    
    Color image & Detected & \multicolumn{2}{c}{Some solutions} & Final result & Final result in 3D \\[-0.05cm]
    & planar segments & \multicolumn{2}{c}{encountered during} & \\[-0.05cm]
    && \multicolumn{2}{c}{the search} &&\\[0.2cm]
    \vspace{-0.6cm}
    \end{tabular}
    }}
    \caption{Overview of our MonteRoom method. From an input color image, we first detect planar regions in the image that can be noisy. Then, by intersecting detected planes, and the frustum planes, we obtain a set of polygonal proposals. Finally, we apply MCTS with refinement to select, and refine, the set of proposals which is consistent with the input image in 2D and 3D.}
    \vspace{-0.5cm}
    \label{fig:method_mr}
\end{figure*}

\subsection{Generating Wall Proposals}

We start by detecting room layout planes, \ie the planes corresponding to the structural components of the scene. For this, we retrain PlaneTR~\cite{tan2021planetr}, which was originally designed for detecting general planes in an input image, to detect the room layout planes instead. More precisely, from an input image $I$ we extract planes $\Pi$. Each detected plane $\pi_i \in \Pi$ is a tuple consisting of several variables: plane normals $n_i$, plane offset $d_i$, and plane embedding vector $e_i$ of size $8$. $n_i$ and $d_i$ are normalized such that $n_i$ is a unit vector. In addition, we use pixelwise embedding $E(I)$ and room layout depth map $D(I)$ in our objective function in Section~\ref{sec:mr_objective} to associate plane embedding vectors with regions in the image and enforce 3D consistency with the scene.

While it is possible to generate a rough room layout estimate already from the detected room layout planes, as we show in Figure~\ref{fig:method_mr}, such approach has several drawbacks. In extreme cases, some of the detected planes might be false positive detections. More frequently, segmented planar regions are inaccurate. Also, planar segmentations are incapable of providing the structural information, \eg connectivities between individual structural elements. Instead, as in~\cite{stekovic2020general}, we intersect all plane triples $\pi_i, \pi_j, \pi_k$, including the frustum planes intersecting the image plane at the boundaries, to obtain candidate room layout corners expressed as  $C^{i,j,k} = (\pi_i, \pi_j, \pi_k)$. We do not consider 3D corners as candidates if their $z$-coordinate is not in range $[0, max(D(I)) + 3]$ as they are very unlikely to correspond to the actual room corner. Then, for each detected plane $\pi_i$ we form a graph consisting of a set of candidate corners $\calC_i$ that lie on $\pi_i$ with edges connecting corners that have exactly $2$ layout planes in common. By finding all cycles in the graphs corresponding to the individual layout planes, we obtain the set of planar polygons $\calP_0$ that we also refer to as wall proposals.   

Instead of directly refining vertices of polygonal proposals as in our MonteFloor implementation, we define only the planes parameters $n$ and $d$ as optimization parameters. Any changes in these parameters directly influence the 2D/3D coordinates of polygon vertices. This is enabled by our design choice of modelling room layout corners as planes triples instead of the actual 2D/3D coordinates.

\subsection{Wall Proposals Selection Game}

In general, we follow the discrete optimization formalization from Stekovic~\etal~\cite{stekovic2020general}. Given an image $I$ as input and a set of planar polygonal proposals $\calP_0$, we want to recover a subset of proposals $\calP \subset \calP_0$ which best describes the room layout of the input scene based on the objective function $\calL(\calP)$. 

Hence, a move in our wall proposals selection game consists of selecting one of the polygonal wall (or floor, ceiling) proposals from the pool. We consider all polygons lying on a single detected plane, to be mutually incompatible. Proposals corresponding to two different planes are mutually incompatible if their polygons intersect. In addition, we ensure that all solutions are feasible layouts. If a proposal contains polygon vertex that lies on another plane, then, for the other plane we only consider compatible proposals that contain the corresponding vertex and disable empty nodes on the corresponding tree levels. Hence, for all layout corners in a valid solution all proposals surrounding the corner will be included as well. 

Method presented in~\cite{stekovic2020general} performs a brute-force search to find the optimal set of wall proposals. In contrast, MCTS is more efficient and is very likely to converge to the same solution. In addition, our objective function is differentiable making our method a very attractive possibility as it allows simultaneous discrete and continuous optimization.

\subsection{Implementing the Scene Loss}
\label{sec:mr_objective}

In case of MonteRoom, we implement $\Ls(I,\calP)$ that evaluates a set of planar polygonal proposals $\calP$ with respect to the input image $I$:

\begin{equation}
    \Ls(I, \calP) = \Ldepth(I, \calP) + \Lemb(I, \calP),
\end{equation}

where $\Ldepth(I, \calP)$ measures the 3D reconstruction loss:

\begin{equation}
    \Ldepth(I, \calP) = \lambda_{d} |D(\calP) - D(I)| + \lambda_{n} |N(\calP) - N(I)|,
\end{equation}

where $D(\calP)$ is a depth map of $\calP$ rendered by our differentiable planar polygons renderer as defined in Section~\ref{sec:mr_diif_rend} and $D(I)$ is a depth map of room layout, estimated from $I$. $N(\calP)$ and $N(I)$ are normal maps extracted from the corresponding depth maps. $\Lemb(I, \calP)$ measures the 2D reconstruction quality:

\begin{equation}
    \Lemb(I, \calP) = \lambda_{e} |E(\calP) - E(I)|,
\end{equation}

where $E(\calP)$ is an embedding map of $\calP$ obtained by, differentiably rendering polygons embeddings as defined in Section~\ref{sec:mr_diif_rend}. In practice, we set loss weights to $\lambda_{d}=0.1$, $\lambda_{n}=0.1$, and $\lambda_{e}=1$ .

\subsection{Differentiable Rendering of Planar Polygons and Polygons Embeddings}
\label{sec:mr_diif_rend}

\textbf{Rendering Depth maps of planar polygons.} It is straight-forward to extend our differentiable winding algorithm for rendering depth maps of planar polygons. More precisely, for a pixel location $m$ and proposal $P$ we render the depth value $R_{3D}(m, P)$:

\begin{equation}
    R_{3D}(m, P) = W(m, \text{Poly}(P)) \frac{-d}{n K^{-1} m_h + \epsilon},
\end{equation}
  
 where W(m,Poly(P)) is the winding value at pixel location $m$ for polygon $\text{Poly}(P)$. The fraction term calculates the depth value with $K$ being the intrinsics matrix, $m_h$ homogeneous representation of $m$, and $\epsilon=10^{-4}$ being a very small value that makes the fraction term more stable. Then, for a selected set of proposals $\calP$, we render the depth map $D(\calP)$ by summing up the individual renderings:
 
 \begin{equation}
    D(\calP) = \sum_{P \in \calP} R_{3D}(P),
\end{equation}

\textbf{Rendering embedding maps of planar polygons.} Similarly, for  $m$ and $P$ we render the embedding value $R_{\text{emb}}(m, P)$: 

\begin{equation}
    R_{\text{emb}}(m, P) = W(m, \text{Poly}(P)) e.
\end{equation}

For given $\calP$, we render the embedding map $E(\calP)$ by summing up the individual renderings: 

 \begin{equation}
    E(\calP) = \sum_{P \in \calP} R_{\text{emb}}(P),
\end{equation}

\subsection{Implementing the Regularization Loss}

Due to our formalization, planar polygonal proposals preserve structure during refinement. This enables us to simplify the $\Lreg$ term:

\begin{equation}
    \Lreg(\calP) = \lambda_{a} \frac{1}{\binom{|P|}{2}}\sum_{(P_i,P_j) \subset P} p(\widehat{n_i, n_j}),
\end{equation}

where $p(\alpha)$ is a soft Manhattan constraint penalizing unlikely angles between individual proposals. In practice we enforce \textit{soft} Manhattan constraints and set $p(\alpha) =$


{\normalsize
\begin{align}
max\left(
\begin{array}{l}
G(|\cos \alpha|, \cos \pi / 6, \sigma), \\
G(|\cos \alpha|, \cos 4.5 \pi / 10, \sigma) \\
\end{array}
\right),
\end{align}
}

where $G$ denotes Gaussian distribution normalized to range $[0,1]$. In practice, we set $\sigma_1=\sigma_2=0.035$. We set loss weight to $\lambda_{a}=0.01$. We emphasize that we can still model non-Manhattan room layouts as our regularization loss does not strictly force wall proposals to be orthogonal or parallel but, rather, it adjusts proposals only in cases when the adjustment does not negatively impact the rest of the objective term. 

\section{MonteRoom Evaluation}
\begin{table}[t]
    \centering \addtolength{\tabcolsep}{6pt}
 \scalebox{0.8}{
    \begin{tabular}{@{}l|c c c c c @{}}
    \toprule
    & PE $\downarrow$ & CE2D $\downarrow$ & CE3D $\downarrow$ & RMSE $\downarrow$ & REL $\downarrow$ \\ 
    
    GeoLayout~\cite{zhang2020geolayout} & 5.24 & 4.36 & 12.82 & 0.46 & 0.11 \\ 
    MonteRoom (Ours) & \textbf{4.21} & \textbf{3.51} & \textbf{11.04} & \textbf{0.40} & \textbf{0.10} \\ 
    \midrule

    \end{tabular}}

    \caption{\textbf{Quantitative evaluation of MonteRoom on the test set of \cite{zhang2020geolayout}} We outperform GeoLayout~\cite{zhang2020geolayout}, the current state-of-the-art method , by a considerable margin on all metrics.}
    \vspace{-1.0cm}
    \label{tab:mr_benchmark}
\end{table}

\newlength{\mrniceresultwidth}
\setlength{\mrniceresultwidth}{0.18\textwidth}
\newlength{\mrniceresultwidthhalf}
\setlength{\mrniceresultwidthhalf}{0.09\textwidth}

\newcommand{\mrniceresult}[1]{

\includegraphics[width=\mrniceresultwidthhalf]{figures/mr_qual/iters/#1_best_iter_1.png} &
\includegraphics[width=\mrniceresultwidthhalf]{figures/mr_qual/iters/#1_best_iter_2.png} &
\vspace{-0.1cm}

\multirow{2}{*}[1cm]{\includegraphics[width=\mrniceresultwidth]{figures/mr_qual/#1_lp_2D_poly_overlay.png}} \bigstrut &
\multirow{2}{*}[1cm]{\includegraphics[width=\mrniceresultwidth]{figures/mr_qual/#1_lpdepth.png}} &
\multirow{2}{*}[1cm]{\includegraphics[width=\mrniceresultwidth]{figures/mr_qual/#1_lgt_2D_poly_overlay.png}} &
\multirow{2}{*}[1cm]{\includegraphics[width=\mrniceresultwidth, height=2cm]{figures/mr_qual/#1_3D.png}} \\

\includegraphics[width=\mrniceresultwidthhalf]{figures/mr_qual/iters/#1_best_iter_3.png} &
\includegraphics[width=\mrniceresultwidthhalf]{figures/mr_qual/iters/#1_best_iter_4.png} \\

}

\begin{figure*}[ht!]
    \centering
    \setlength{\tabcolsep}{1pt}
    \scalebox{1.0}{
    \scalebox{1}{
    \begin{tabular}{cccccc}
    \mrniceresult{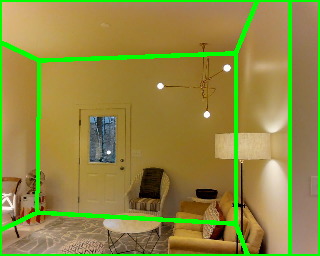}
    \mrniceresult{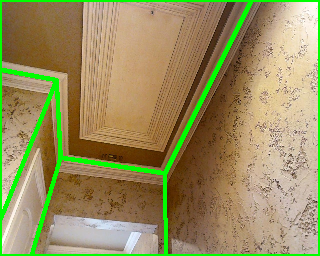}
    \mrniceresult{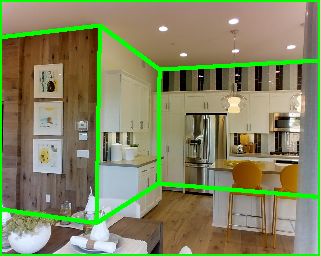}
    \mrniceresult{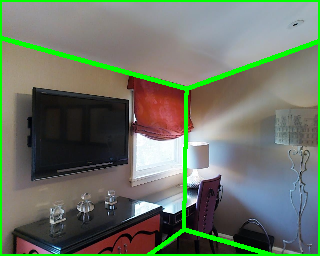}
    \mrniceresult{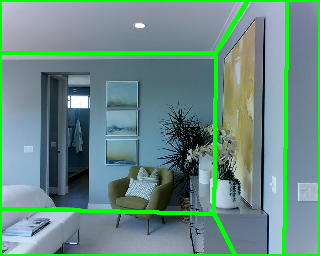}
    \mrniceresult{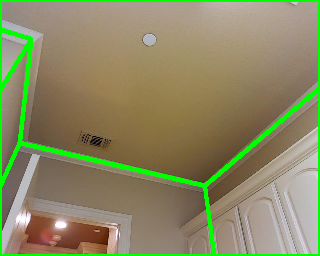}
    \mrniceresult{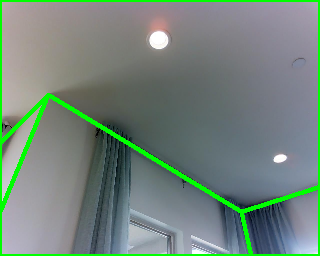}

    \multicolumn{2}{c}{Alternative solutions} & Our 2D reconstruction & Our depth map & G.t. 2D reconstruction  & \hspace{0.1cm} Our 3D reconstruction 

    \end{tabular}}}
    \caption{\textbf{Room layout reconstructions.} Our MonteRoom approach generalizes to a range of room layout configurations including non-cuboid and non-Manhattan layouts (example in fourth row). We observe that MonteRoom identifies correct solutions among a variety of alternatives. Among the four visualized alternatives in first column, the fourth solution is already the best choice among all alternatives but we observe that final result aligns better with the input scene because of our refinement procedure. In addition, we observe that the structure of room layout remains in-tact during refinement as we only optimize polygons indirectly through planes parameters, in contrast to our MonteFloor application that directly refines polygon vertices.}
    \label{fig:mr_qual}
\end{figure*}

In this section, we evaluate our MonteRoom method on the Matterport3D-Layout dataset~\cite{zhang2020geolayout}. The dataset consists of $4939$ training images, $456$ validation images, and $1965$ test images. The training and the validation color images contain manual annotations for 3D room layouts and the corresponding planes parameters. For the test set, the ground truth is not available, and the results are submitted and evaluated directly by the authors of the benchmark. We train PlaneTR~\cite{tan2021planetr} on the training set for $60$ epochs and reduce the maximum number of planes to $10$ as we observed that none of the scenes in the dataset exceeds this number. 

As the original Matterport3D-Layout benchmark provides many metrics for evaluating the quality of room layout depth maps, we show only the Root Mean Square Error (RMSE) and the absolute RELative difference (REL), as we did not find other depth map related  metrics to be more informative in our evaluations. In addition, the benchmark measures three additional metrics: Pixel Error (PE) measures the accuracy of 2D room layout segmentation, and 2D Corner Error (CE2D) and 3D Corner Error (CE3D) measure the distance of predicted room layout corners to their actual locations. In addition, CE2D and CE3D strongly penalize solutions where the number of predicted corners is different than the number of ground truth corners.  

We compare MonteFloor to the GeoLayout baseline~\cite{zhang2020geolayout} quantitatively in Table~\ref{tab:mr_benchmark} and observe significant improvements. Our MonteRoom approach outperforms the GeoLayout baseline on all metrics. GeoLayout greedily assigns plane parameters based on their clustering post-processing mechanism. Instead, our MonteRoom is able to focus on the most promising solutions during the search and at the same time refine selected solutions to improve the fitness in respect to the input scene.

In addition, we provide qualitative results in Figure~\ref{fig:mr_qual}. Our MonteRoom approach is able to consistently select configurations that maximize fitness to the input scenes. We do not apply any hard constraints, except for planarity, which makes our approach generalizable to a range of non-cuboid and non-Manhattan room layout configurations. In particular, we demonstrate that our approach is able to distinguish between different configurations and refine the most promising solution.

\begin{table}[t]
    \centering \addtolength{\tabcolsep}{6pt}
 \scalebox{0.76}{
    \begin{tabular}{@{}l|c c c c c @{}}
    \toprule
    & PE $\downarrow$ & CE2D $\downarrow$ & CE3D $\downarrow$ & RMSE $\downarrow$ & REL $\downarrow$ \\ 
    planes only & 5.53 & - & - & 0.52 & 0.14 \\ 
    no refin. step & 5.45 & 2.32 & 11.60 & \textbf{0.46} & \textbf{0.12} \\ 
    w/o $\Ldepth$ & 4.72 & \textbf{2.25} & 11.85 & 0.47 & 0.13 \\
    w/o $\Lreg$ & 4.61 & 2.40 & 11.84 & \textbf{0.46} & \textbf{0.12} \\
    w/o $\Lemb$ & 6.16 & 3.80 & 13.41 & 0.52 & 0.13 \\

    complete & \textbf{4.56} & \textbf{2.25} & \textbf{11.47} & \textbf{0.46} & \textbf{0.12} \\ 
    \midrule

    \end{tabular}}

    \caption{\textbf{Ablation study on the validation set of~\cite{zhang2020geolayout}}. The plane detection module alone cannot achieve the same performance levels and that the module does not model room layout corners. Removing any of the objective terms and not relying on the refinement step have negative impacts on metrics. Our method shows largest dependence on the $\Lemb$ while we believe that other terms would be more significant in practical applications.}
    \vspace{-0.8cm}
    \label{tab:mr_ablation}
\end{table}

\textbf{Ablation Study.} In our ablation study in Table~\ref{tab:mr_ablation} we evaluate our adaptation of PlaneTR~\cite{tan2021planetr} network, and the influence of our refinement step and individual objective terms on the final result. We perform these evaluations on the validation set of Matterport3D-Layout. From the outputs of PlaneTR we can generate planar segments by associating planar and pixelwise embeddings, however even though the depth map metrics are relatively good, the PE metric is still high and such approach does not model room layout corners. For the ablation study without refinement we reduce the number of steps to $100$ as we observed that the search converges to a good solution very fast. However, because of imperfect initialization we observe that the refinement is indeed needed, which is best observable through significant improvement for the PE metric. Both the 3D reconstruction loss and the regularization loss mostly show their contribution in terms of 3D corner error. While significance of these terms does appear to be minor in our experiments, we do believe that these terms could still be very useful in practice. Due to high quality of planar embeddings, our optimization relies significantly on the term measuring the 2D reconstruction quality. Removing this term results in a significant drop in performance of our method. 

We also measure the computation time of our method. Interestingly, without refinement, our search is very fast with run time of $1-2$ seconds. However, this run time could be easily improved by pre-rendering proposals like in~\cite{MCSS} which is the bottleneck in objective term calculation. With refinement our method converges in roughly $1$ minute. Hence, in practice for some applications, we could improve run time by performing refinement only in later iterations of MCTS.

\section{Conclusion}

We have shown how MCTS with refinement step can be applied to problems in scene understanding that can be formalized as proposals selection games. We applied our method to two problems of high relevance in AEC, namely floor plan estimation and 3D room layout estimation, and demonstrated high accuracy in both of these applications. 

Beyond these applications, we believe our approach is general. All that is needed to adapt it to other scene understanding problems is (1) a way to generate proposals and (2) a differentiable function to evaluate, and optimize, the quality of a solution.  We hope our work will inspire researchers to consider  problems with complex interactions between objects and obtain robust and accurate results. 



%

\ifCLASSOPTIONcompsoc
  \section*{Acknowledgments}
\else
  \section*{Acknowledgment}
\fi

This work was supported by the Christian Doppler Laboratory for Semantic 3D Computer Vision, funded in part by Qualcomm Inc.

\ifCLASSOPTIONcaptionsoff
  \newpage
\fi




~\newpage~\newpage
\bibliographystyle{IEEEtran}
\bibliography{string,0bib}

%



%

\begin{IEEEbiography}[{\includegraphics[width=1in,height=1.25in,clip,keepaspectratio]{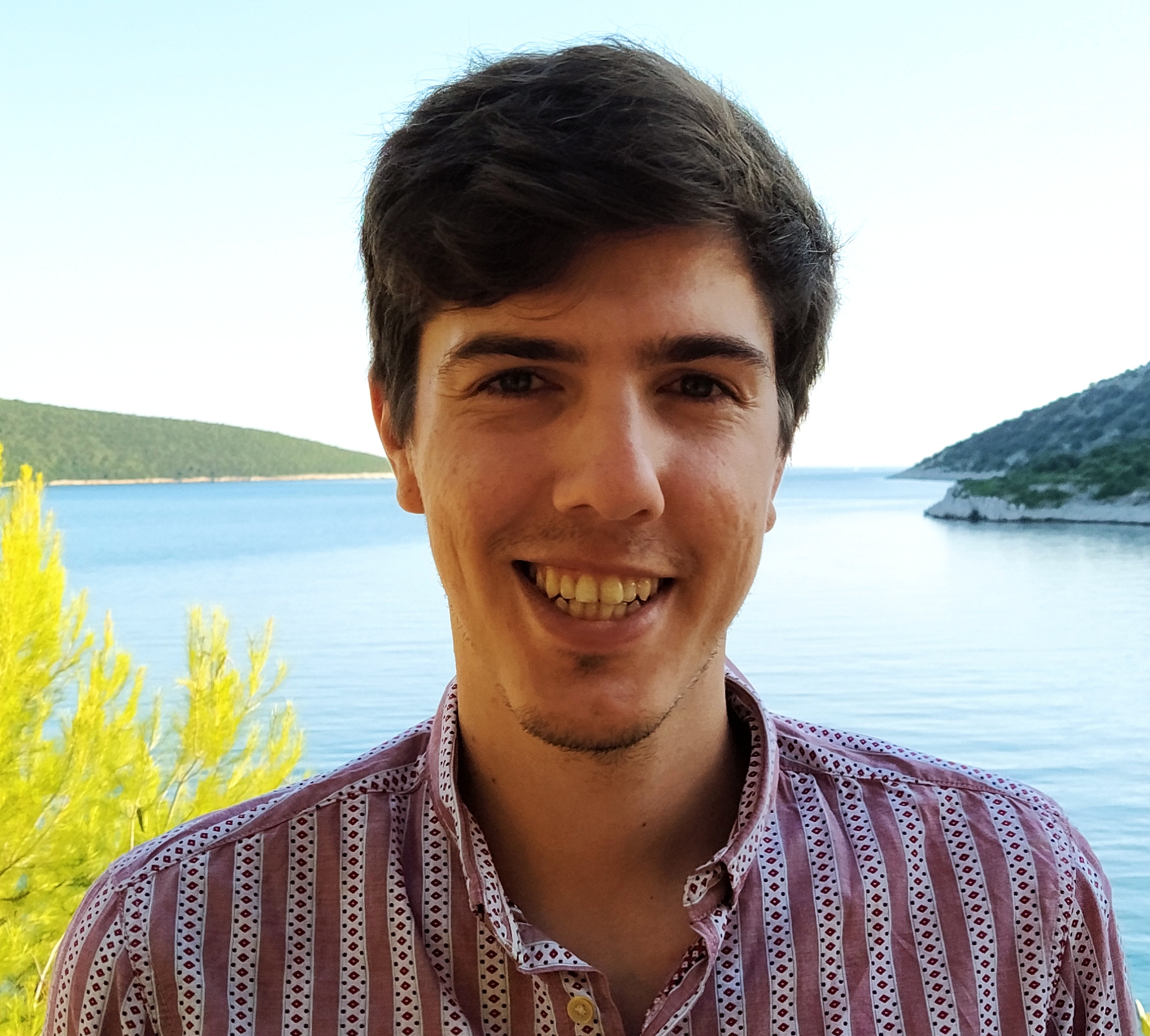}}]{Sinisa Stekovic}
is a Project Assistant~(Ph.D.) at the Institute of Computer Graphics and  Vision at Graz University of Technology (TUG) under the supervision of Associate Prof. Friedrich Fraundorfer and Prof. Vincent Lepetit. He graduated from Graz University of Technology as M.Sc.~(Dipl.-Ing.) in Computer Science in 2018 with distinction. His current research focus lies in applying Machine Learning and Computer Vision methods in the field of 3D scene understanding, mostly focusing on 3D reconstruction of structural elements in indoor environments. 
\end{IEEEbiography}


\vspace{-1.0cm}
\begin{IEEEbiography}[{\includegraphics[width=1in,height=1.25in,clip,keepaspectratio]{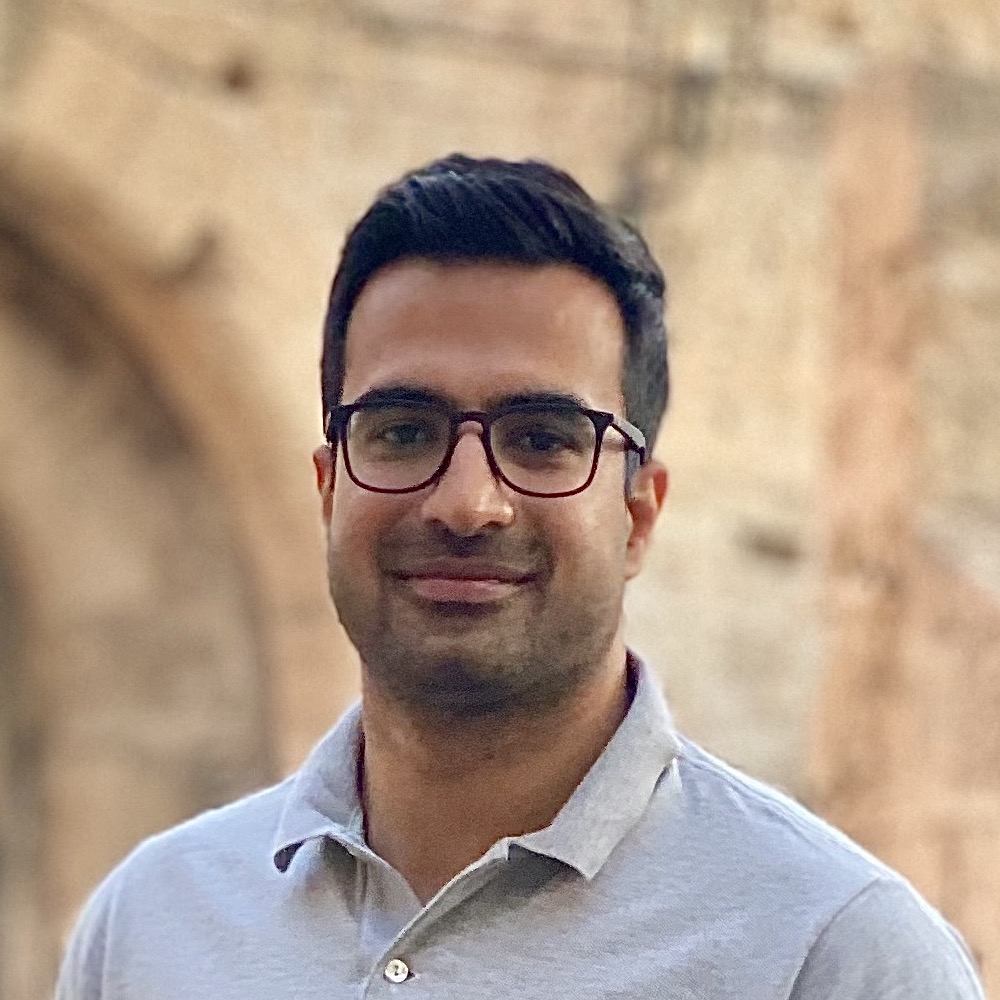}}]{Mahdi Rad }
is a postdoctoral scholar at the Institute of Computer Graphics and Vision of Graz University of Technology (TUG). He received his Ph.D. degree with distinction from TUG, under the supervision of Prof. Vincent Lepetit, in 2018. Before that, he obtained his B.Sc. and M.Sc. degrees in Computer Science from EPFL, Switzerland, in 2012 and 2014. His research interests include machine learning methods for computer vision, specifically methods for 3D scene understanding, 3D object pose estimation, 3D hand pose estimation, hand-object interaction, semi supervised learning, and domain adaptation.
\end{IEEEbiography}


\begin{IEEEbiography}[{\includegraphics[width=1in,height=1.25in,clip,keepaspectratio]{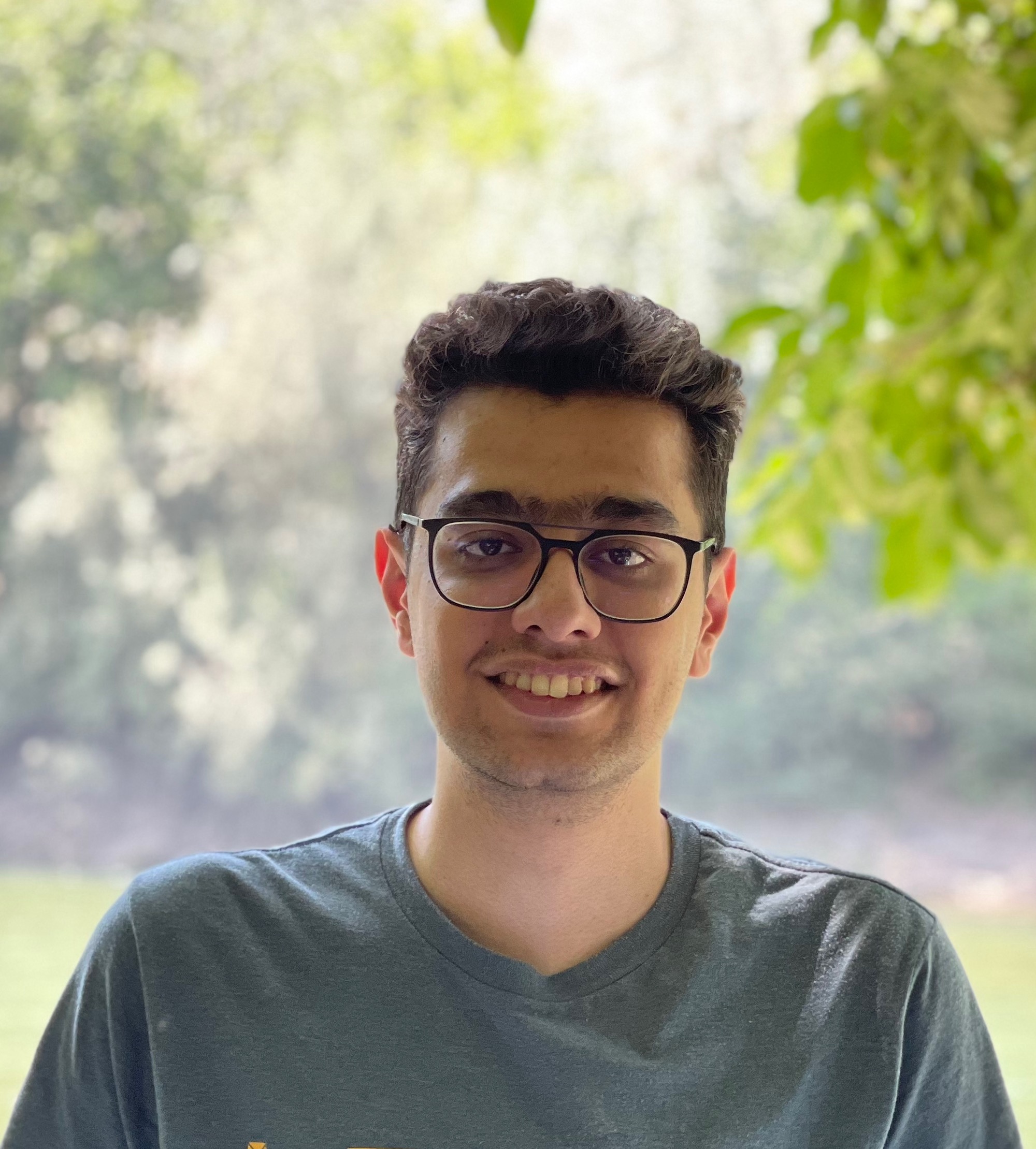}}]{Alireza Moradi}
is a research intern at the Institute of Computer Graphics and Vision at Graz University of Technology (TUG), under the supervision of Prof. Vincent Lepetit. He is also a Computer Engineering undergraduate student at Iran University of Science and Technology (IUST), Tehran, and is about to graduate. His research interests include developing Machine Learning methods for Computer Vision, with accent on 3D Vision problems.
\end{IEEEbiography}

\newpage

~\vspace{-4.0cm}
\begin{IEEEbiography}[{\includegraphics[width=1in,height=1.25in,clip,keepaspectratio]{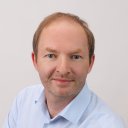}}]{Friedrich Fraundorfer}
received the PhD degree in Computer Science from TU Graz in 2006. He is an assistant professor at the Graz University of Technology, Austria. He had post-doc stays at the University of Kentucky, at the University of North Carolina at Chapel Hill, at ETH Zürich and acted as the deputy director of the Chair of Remote Sensing Technology at the Technische Universität München from 2012 to 2014. His main research areas are 3D computer vision and robot vision.
\end{IEEEbiography}

~\vspace{-7.5cm}
\begin{IEEEbiography}[{\includegraphics[width=1in,height=1.25in,clip,keepaspectratio]{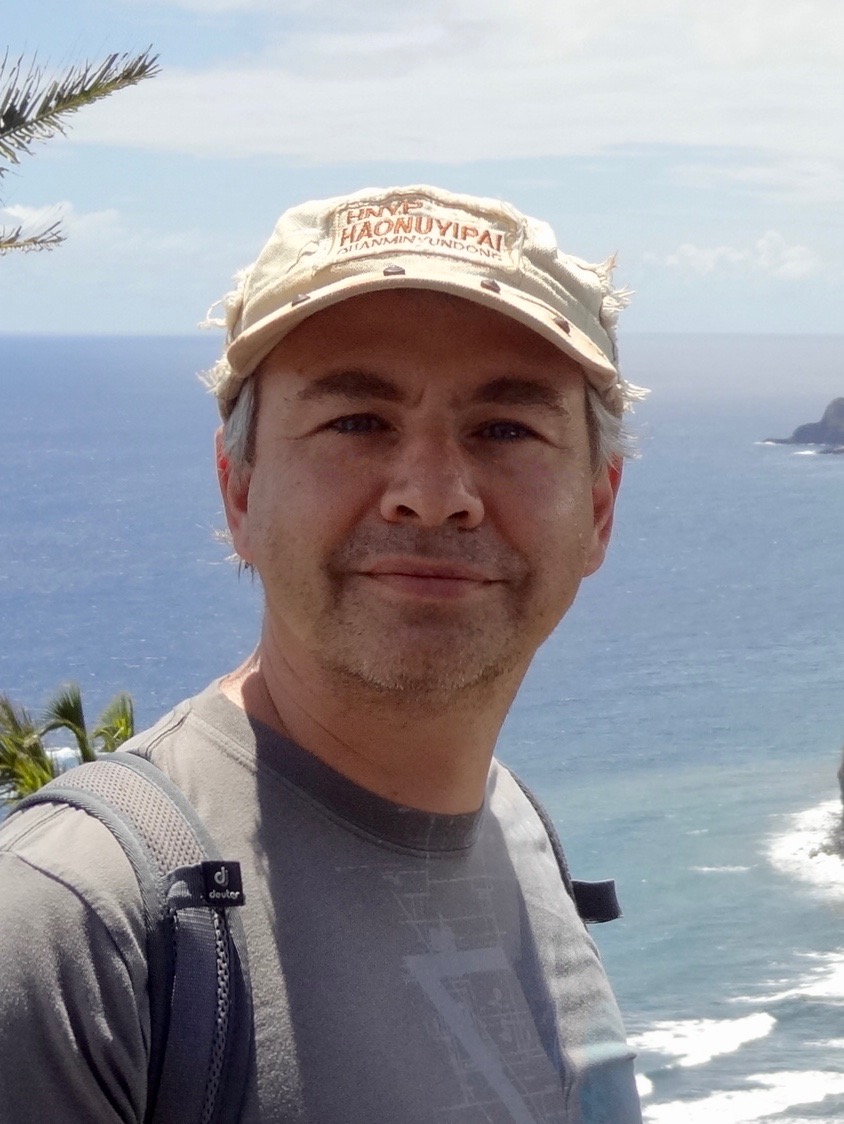}}]{Vincent Lepetit}
is a director of research at ENPC ParisTech, France, since 2019. Prior to being at ENPC, he was a full professor at the Institute for Computer Graphics and Vision, Graz University of Technology, Austria, where he still supervises a research group, and before that, a senior researcher at the Computer Vision Laboratory (CVLab) of EPFL, Switzerland. His research interest are at the interface between Machine Learning and 3D Computer Vision, with current focus on 3D scene understanding from images. He often serves as an area chair for the major computer vision conferences (CVPR, ICCV, ECCV) and is an editor for PAMI, IJCV, and CVIU. 
\end{IEEEbiography}




\end{document}